\documentclass{article}

\usepackage{amsmath,amsfonts,bm}

\def\eqref#1{equation~\ref{#1}}

\def\1{\bm{1}}

\DeclareMathAlphabet{\mathsfit}{\encodingdefault}{\sfdefault}{m}{sl}
\SetMathAlphabet{\mathsfit}{bold}{\encodingdefault}{\sfdefault}{bx}{n}

\newcommand{\E}{\mathbb{E}}

\DeclareMathOperator*{\argmax}{arg\,max}

\usepackage{microtype}
\usepackage{graphicx}
\usepackage{booktabs} 
\usepackage[flushleft]{threeparttable}
\usepackage{array,booktabs,makecell}
\usepackage{hyperref}

\usepackage[accepted]{icml2023}

\usepackage{amsmath}
\usepackage{amssymb}
\usepackage{mathtools}
\usepackage{amsthm}

\usepackage[capitalize,noabbrev]{cleveref}

\theoremstyle{plain}
\newtheorem{theorem}{Theorem}[section]

\newtheorem{lemma}[theorem]{Lemma}

\theoremstyle{definition}

\theoremstyle{remark}
\newtheorem{remark}[theorem]{Remark}

\usepackage[textsize=tiny]{todonotes}

\usepackage{url}
\usepackage{hyperref}       
\usepackage{xcolor}
\usepackage{amsthm}
\usepackage{chngcntr}
\usepackage{apptools}

\usepackage{graphicx}
\usepackage{graphics}
\usepackage{parskip}
\usepackage{amsmath}
\usepackage{subfig}
\usepackage{dsfont}
\usepackage{natbib}
\usepackage[leftcaption]{sidecap}
\usepackage{wrapfig}
\usepackage[normalem]{ulem}
\usepackage{soul}
\usepackage{forest}
\useforestlibrary{edges}
\usepackage[nopar]{lipsum}

\icmltitlerunning{Pareto Regret Analyses in Multi-objective Multi-armed Bandit}

\begin{document}

\twocolumn[
\icmltitle{Pareto Regret Analyses in Multi-objective Multi-armed Bandit}



\icmlsetsymbol{equal}{*}

\begin{icmlauthorlist}
\icmlauthor{Mengfan Xu}{yyy}
\icmlauthor{Diego Klabjan}{yyy}
\end{icmlauthorlist}

\icmlaffiliation{yyy}{Department of Industrial Engineering and Management Sciences, Northwestern University, Evanston, IL 60208, U.S.A.}

\icmlcorrespondingauthor{Mengfan Xu}{MengfanXu2023@u.northwestern.edu}
\icmlcorrespondingauthor{Diego Klabjan}{d-klabjan@northwestern.edu}

\icmlkeywords{Machine Learning, ICML}

\vskip 0.3in
]



\printAffiliationsAndNotice{}  

\begin{abstract}
We study Pareto optimality in multi-objective multi-armed bandit by providing a formulation of adversarial multi-objective multi-armed bandit and defining its Pareto regrets that can be applied to both stochastic and adversarial settings. The regrets do not rely on any scalarization functions and reflect Pareto optimality compared to scalarized regrets. We also present new algorithms assuming both with and without prior information of the multi-objective multi-armed bandit setting. The algorithms 
are shown optimal in adversarial settings and nearly optimal up to a logarithmic factor in stochastic settings simultaneously by our established upper bounds and lower bounds on Pareto regrets. Moreover, the lower bound analyses show that the new regrets are consistent with the existing Pareto regret for stochastic settings and extend an adversarial attack mechanism from bandit to the multi-objective one.
\end{abstract}

\section{Introduction}

Multi-armed bandit (MAB) is a sequential paradigm where players choose arms and receive reward values from an environment at each time step. It usually aims at maximizing the cumulative reward of the player, or equivalently minimizing the regret formulated as the difference between the rewards of the best arm and the obtained rewards of the player. Multi-objective Multi-armed bandit (MO-MAB) extends reward values of arms to multi-dimensional reward vectors and thereby changing the nature of the MAB problem significantly as a result of the complicated order relationships of vectors. Rewards can be optimal in one dimension but sub-optimal in other dimensions, leading to multiple optimal arms. Formally, given time horizon $T$, at time step $t\leq T$ the player chooses one arm $a_t$ among $K$ arms and receives the $D$-dimensional vector $r^{a_t,t}$ among rewards $(r^{i,t})_{i=1}^{K}$. The goal is to optimize the total reward vector $\sum_{t=1}^T r^{a_t,t}$ by minimizing some regret metric measuring how far the player is away from optimality. 

There are two ways to define optimality: Pareto optimality in the reward vector space and Scalarized optimality by scalarizing reward vectors. Pareto optimality admits a Pareto optimal front defined as the set of rewards of optimal arms determined by the Pareto order relationship. With limited information based on the definition of MO-MAB, it is a great challenge to directly estimate the Pareto optimal front while crucial for an optimal strategy. Scalarization is often used to transform reward vectors to scalars and thereby reducing MO-MAB to MAB depending on scalarization functions, which, however, has the following several limitations. First, how to properly choose such scalarization functions may be troublesome. Linear and Chebyshev options have been discussed in~\citep{drugan2013designing}, despite of the fact that they may not fully explore non-convex Pareto optimal fronts. Secondly, algorithms can be less robust since their performance highly depends on the specific scalarization functions. In other words, what is optimal for a scalarization function, can result in large scalarized regrets for other functions. Therefore, Pareto optimality in the reward vector space makes more sense and thereby being the main focus herein.  

MAB usually falls into the category of either stochastic MAB or adversarial MAB depending on how rewards are generated. Following similar assumptions on the reward process, we consider MO-MAB as stochastic MO-MAB and adversarial MO-MAB, respectively. In stochastic MO-MAB, the rewards of each arm at different time steps are assumed to be i.i.d.,  following a multi-dimensional distribution of which the mean vector is time-invariant. Henceforth, the Pareto optimal set is the set of arms with mean vectors that are Pareto optimal and the Pareto optimal reward set (front) is the set of corresponding mean vectors, which are constant over time. A regret metric concerning Pareto optimality, namely Pareto regret, is then defined as the cumulative distance between the expected reward received at each time step and the Pareto optimal front~\citep{drugan2013designing}. However, this regret measure cannot generalize to adversarial MO-MAB by noting that the Pareto optimal front is ill-defined. Precisely, for adversarial MO-MAB, the rewards are arbitrarily specified by adversaries at each time step and thereby varying with time and even depending on past observations. Meanwhile, when the adversary generates rewards based on a distribution depending on the state (context) at each time step, i.e. a context-dependent reward generator, it boils down to contextual MO-MAB. It necessitates a new way to determine Pareto optimality, Pareto optimal front and subsequently a Pareto regret measure in adversarial settings, which has not yet been studied.

Traditionally, algorithms designed for MO-MAB work either by modifying the methods for MAB (MAB-based) or by importing the mechanisms from Multi-objective Optimization (MOO) (MOO-based). More specifically, MAB-based algorithms alternate the estimations of the best arm by approximating a Pareto optimal set, from which an arm is randomly selected. Examples include Pareto UCB~\citep{drugan2013designing} as optimistic in face of uncertainty and MOSS++~\citep{zhu2020regret} as an extension of MOSS~\citep{audibert2009minimax}. Their formulations and analyses are limited to stochastic MO-MAB.  
While a lot of attention has been given to algorithms for variants of stochastic MO-MAB~(see \citep{huyuk2021multi,van2014multi}), a counterpart remains unexplored in adversarial settings, including the lack of a valid formulation and analyses which we fully cover herein. Moreover, the aforementioned MAB-based algorithms may result in linear Pareto regret when applied to our defined adversarial settings. Precisely, we provide a regret analysis of Pareto UCB in the adversarial regime where Pareto UCB suggests a linear regret.

Some MOO-based approaches adapt Multi-objective Evolutionary Algorithms~\citep{hong2021evolutionary} to bandit problems and ~\citep{yahyaa2014annealing} propose the annealing knowledge gradient descent method, both of which, however, are examined empirically without theoretical guarantees. Analyses for MOO under uncertainty provide possibilities for adversarial MO-MAB given that the objective functions can be black-box. Recently, a minimax regret formulation has been suggested by~\citep{groetzner2022multiobjective}, though Pareto optimality is again neither defined nor potentially guaranteed by proposing algorithms with an optimal minimax regret. 
An algorithm that guarantees theoretically optimal Pareto regret for adversarial MO-MAB has not yet been proposed, let alone achieving optimality for both adversarial and stochastic MO-MAB. 

Herein we propose a formulation of the adversarial MO-MAB problem where rewards are determined arbitrarily under no assumptions on their distributions. Moreover, we formally introduce its Pareto optimal front and Pareto regret and Pareto pseudo regret, which makes it a possibility to consider algorithms from a perspective of Pareto optimality. To our best knowledge, this is the first work formally introducing the general MO-MAB setting and considering the task of Pareto regret optimization in adversarial MO-MAB, in line with both MAB and stochastic MO-MAB. Surprisingly, we can extend these regrets to stochastic settings without any modifications. On the basis of existing literature on stochastic MO-MAB, the achievements of this paper are to build the theoretical connection between general MO-MAB and MAB which allows us to characterize Pareto regret generally, and to develop theoretically effective methods that can achieve Pareto optimality in both stochastic and adversarial MO-MAB. 

In this paper, we characterize the Pareto regret explicitly and propose algorithms that achieve regrets of order $\sqrt{T}$ in adversarial MO-MAB and regrets of order $\log{T}$ or $(\log{T})^2$ in stochastic MO-MAB. More specifically, when given whether it is a stochastic schema or adversarial schema before the game starts, our algorithm MO-KS, which switches between the UCB algorithm and the EXP3.P algorithm, obtains Pareto regret of the same orders as the regret in MAB in the sense that the regret is of order $\log{T}$ and $\sqrt{T}$ in stochastic MO-MAB and adversarial MO-MAB, respectively. Otherwise, when this prior information is not available a different algorithm MO-US is proposed herein that combines the EXP3 and UCB algorithms, as a modification of the existing IEXP3++ algorithm for MAB~\citep{seldin2017improved}. Consequently, its Pareto pseudo regret has an order of $(\log{T})^2$ in stochastic MO-MAB and $\sqrt{T}$ in adversarial MO-MAB, respectively, without necessarily knowing the settings of MO-MAB.

To claim optimality of our proposed algorithms and to provide the worst-case scenario analyses for the proposed framework, we are the first to provide regret lower bound analyses for both stochastic and adversarial MO-MAB. More precisely, we show that in stochastic MO-MAB, the expected Pareto regret and the Pareto pseudo regret cannot be smaller than order $\log{T}$ whatever algorithms are deployed. For adversarial MO-MAB, a regret of order greater than or equal to $\sqrt{T}$ results from any randomized algorithm.  The lower bounds on Pareto regret match the upper bounds of MO-KS in that they have the same order with respect to $T$, which implies that it is optimal in both settings. Note that for MO-US, the lower bound coincides with the upper bound in adversarial settings and is almost the same as the upper bound in stochastic ones up to the $\log{T}$ multiplicative factor, i.e. MO-US is optimal in adversarial MO-MAB and nearly optimal up to a $\log T$ factor in stochastic MO-MAB. 
Moreover, we derive the lower bound of Pareto UCB, a specific algorithm that is optimal in stochastic MO-MAB, by constructing instances in adversarial MO-MAB. We report that Pareto UCB can result in a linear Pareto regret far away from optimal, which further necessitates the proposed algorithms. 
As a by-product, the instances form a generalization of an adversarial attack on UCB in MAB~\citep{jun2018adversarial}. We provide an attack cost analysis and pre- and post-attack regret analysis of Pareto UCB for MO-MAB herein.

The rest of the paper is as follows. In Section~\ref{sec:lr}, we present the existing work in the domain of MO-MAB. Then in Section~\ref{sec:pre}, we introduce the preliminaries including notations and formulations in the context of MO-MAB. We proceed by proposing an algorithm with optimality in both stochastic and adversarial MO-MAB, as well as a nearly optimal one up to a $\log T$ factor, and establishing their regret upper bounds in Section~\ref{sec:ub}. Finally, in Section~\ref{sec:lb}, we provide full analyses on the regret lower bound and on a special case when Pareto UCB is adopted. 

\section{Literature Review}\label{sec:lr}

MAB is well understood and many algorithms have proven to be effective for MAB both theoretically and numerically. For stochastic MAB, UCB and Thompson Sampling achieve optimal regret of order $\log{T}$. For adversarial MAB, \citep{auer2002nonstochastic} propose EXP3 with an optimal regret $\sqrt{T}$. Moreover, in the work of~\citep{seldin2014one}, EXP3++ algorithm is designed as a nearly optimal solution to both stochastic and adversarial setting, up to a $\log T$ factor. More specifically, the EXP3++ algorithm dynamically updates two levers used for EXP3 and UCB, respectively, by estimating the assumed nature of rewards, and yields a sample rule accordingly that is a weighted average of EXP3 and UCB. The SAO algorithm~\citep{auer2016algorithm} improves it by having smaller regret in the stochastic setting but with a larger regret in the adversarial setting instead and with more complicated steps. \citep{seldin2017improved} improve on EXP3++ by a careful parameterization that leads to smaller regret and does not require a prior $T$ and~\citep{zimmert2019optimal} finally arrive at optimum for both settings under certainly strong conditions. MO-MAB extends the single-dimensional objectives of MAB to be multi-dimensional ones and makes it strikingly challenging to develop optimal methods. To this end, it is natural to extend MAB algorithms to MO-MAB and Pareto UCB is such a success. Specifically, Pareto UCB ~\citep{drugan2013designing} constructs upper bounds of confidence intervals for each arm and determines a Pareto optimal front instead of the best arm in the presence of Pareto optimality. An arm is randomly pulled from the set. Pareto regret of order $\log{T}$ for Pareto UCB is guaranteed by the multi-dimensional Chernoff-Hoeffding bound. Further improvement on it can be found in ~\citep{drugan2014pareto}. However, they only work for stochastic MO-MAB. There are variants of stochastic MO-MAB, such as PF-LEX for MO-MAB with a lexicographical order and satisficing objectives (which is a relaxation of maximizing that allows for suboptimal less risky decision making in the face of uncertainty) ~\citep{huyuk2021multi} and MO-linUCB for contextual MO-MAB~\citep{mehrotra2020bandit}. But no attention has been given to the framework of adversarial MO-MAB.

Another line of work on MO-MAB is to modify Multi-objective Optimization (MOO) methods. A lot of algorithms have shown empirically great performance in real applications, such as larg-scale MOEA using evolutionary algorithms dealing with the classic trade-off between exploration and exploitation~\citep{hong2021evolutionary} and Annealing Pareto Knowledge Gradient as a variant of Thompson Sampling~\citep{yahyaa2014annealing}. But the lack of a theoretical guarantee presents a concern since data-driven methods may be less robust and more sensitive, which motivates focus on theoretically effective algorithms. MOO methods consider both deterministic settings and settings under uncertainty, where the former can be adapted to stochastic MO-MAB with deterministic mean vectors and the latter raises the possibility of use in adversarial MO-MAB. For stochastic MO-MAB, various scalarization techniques from MOO are proposed for dimension reduction, such as Linear and Chebyshev, with which the problem essentially degenerates to MAB. But scalarizations may result in a loss of information and be highly problem-dependent as stated earlier. Optimizing the General Gini Index (GGI) using online convex optimization~\citep{busa2017multi} extends the concept of the Gini Index by means of ordered weighted average, albeit the regret metric is again defined on GGI. 
For MOO under uncertainty, utility functions including both linear and non-linear options are often applied in economics, whereas they work similarly as scalarizarion methods. Recently, \citep{groetzner2022multiobjective} suggest a relative minimax regret approach with theoretical guarantees. Specifically, the minimax regret first takes maximum over the random variables for each dimension and then minimum over the decision variables using Pareto order relationships. However, it 
requires precomputing all optimal values with respect to random variables, which does not generalize to on-the-fly bandit settings. 
Herein we fill the gap by formally introducing the Pareto regret for adversarial MO-MAB and by proposing algorithms that show optimality or near optimality in both stochastic and adversarial MO-MAB.

Lower bound analyses on regret play an important role in studying the optimality of algorithms in MO-MAB. On the one hand, the work on MO-MAB algorithms usually shows lower bounds on regret depending on the problem settings and what regret metrics are used. For Pareto regret, \citep{drugan2013designing} argue a lower bound of order $\log{T}$ for stochastic MO-MAB that matches both the regret upper bound of Pareto UCB and the regret lower bound for stochastic MAB. For our newly proposed Pareto regrets, we also establish their lower bounds in both stochastic and adversarial MO-MAB, which are consistent not only with most of the upper bounds, but with~\citep{drugan2013designing} in stochastic settings, which further supports the validity of the proposed regret metrics. On the flip side, an adversarial attack on UCB~\citep{jun2018adversarial} shows that UCB may suffer a linear regret given a small attack cost, while the vulnerability of Pareto UCB is not yet studied which potentially provides bad cases where Pareto UCB does not work. In this paper, we consider the adversarial attack on Pareto UCB, which not only justifies its use in MO-MAB, but as an evidence of the limitation of Pareto UCB for adversarial MO-MAB.

\section{Preliminaries}\label{sec:pre}

In this section, we start with the notations used throughout the paper and then formally introduce Pareto regret including the existing one for stochastic MO-MAB and our newly proposed ones that capture both stochastic MO-MAB and adversarial MO-MAB. The regret analyses are presented in the next section.

\subsection{Notation}

For vectors, a Pareto order relationship has been introduced in~\citep{drugan2013designing} and ~\citep{drugan2014pareto}. There are several possible order relationships between two vectors $a$ and $b$. Vector $a$ is said to weakly dominate $b$, written as $a \succeq b$ or  $b \preceq a$, if and only if for any dimension $d$, $a_d \geq b_d$. Removing the equality for at least one dimension gives us dominating, which is denoted by $a \succ b$ or $b \prec a$.  Vector $a$ is incomparable with $b$, i.e. $a || b$, if there exists a dimension $d$ such that $a_d > b_d$ and another dimension $d^{\prime}$ satisfying $a_{d^{\prime}} < b_{d^{\prime}}$. We say that vector $a$ is non-dominated by $b$ if there is dimension $d$ such that $a_d > b_d$.

We consider MO-MAB with $K$ arms being $\{1,2,\ldots,K\}$ and $D$-dimensional rewards and time horizon $T$. The chosen arm by a player at each time step $t$ is denoted by $a_t$ and only the reward of $a_t$, namely $r^{a_t,t}$, is revealed to the player. Value $N_i(t)$ is the total number of pulls of arm $i$ by the player up to time $t$. In the stochastic setting, at each time step $t$, the reward $r^{i,t} = (r^{i,t}_l)_{1 \leq l \leq D}$ of arm $i$ follows a distribution with time-invariant mean vector $\mu^{i} = (\mu^{i}_l)_{1 \leq l \leq D}$ satisfying $1 \succeq \mu^{i} \succeq 0$. For adversarial MO-MAB, the reward $0 \preceq r^{i,t} \preceq 1$ of each arm $i$ is specified by an adversary which can be adaptive or oblivious where the former depends on past actions of players and the latter does not.

\subsection{Pareto Optimality and Regret}

\subsubsection{Optimality}

The Pareto order relationship can determine the Pareto optimality of arms as follows. Formally, in stochastic settings, an arm $j$ is said to be Pareto optimal, if $\mu^{j}$ is non-dominated by the reward mean vectors of any other arm and the set of Pareto optimal arms is named as the Pareto optimal set, denoted by $O_A$~\citep{drugan2013designing}. We propose to define the Pareto optimal set $O^{\prime}_A$ and $\bar{O}^{\prime}_A$ based on Pareto optimality with respect to the cumulative reward vector $\sum_{t=1}^Tr^{i,t}$ and $\sum_{t=1}^TE[r^{i,t}]$, respectively, for both stochastic settings and adversarial settings. 
Formally, $O^{\prime}_A = \{i^*:\sum_{t=1}^Tr^{i^*,t} \text{ is non-dominated by } \sum_{t=1}^Tr^{j,t}, \text{ for any arm } \newline j \neq i^*\}, 
\bar{O}^{\prime}_A = \{i^*:\sum_{t=1}^TE[r^{i^*,t}] \text{ is non-dominated by }$ \newline $\sum_{t=1}^TE[r^{j,t}] \text{ for any arm } j \neq i^*\} $  allow us to define a uniformly effective Pareto regret for MO-MAB. Note that $\bar{O}^{\prime}_A$ is actually the same as $O_A$ in stochastic settings. We denote the corresponding Pareto optimal front as $O = \{\mu^a:a \in O_A\}$, $O^{\prime} = \{\sum_{t=1}^Tr^{a,t}:a \in O^{\prime}_A\}$, and $\bar{O}^{\prime}  = \{\sum_{t=1}^T\E[r^{a,t}]: a \in \bar{O}^{\prime}_A \}$.

\subsubsection{Regret}

We now proceed to introduce Pareto regrets by measuring the distance between the obtained rewards and the rewards of arms in the Pareto optimal set with a metric, consistent with in~\citep{drugan2013designing}. In~\citep{drugan2013designing}, Pareto regret for Stochastic MO-MAB is denoted by $R_T = \sum_{t=1}^{T}Dist(\mu^{a_t}, O) = \sum_{i=1}^{K}N_i(T)\cdot Dist(\mu^i , O)$,
where the distance measure $Dist(a,O)$ between a vector $a \preceq \sigma$ for every $\sigma \in O$ and a set $O$ is defined by $Dist(a,O) = \min_{\epsilon \geq 0}\{\epsilon: a +\epsilon1 || \sigma \text{ for every } \sigma \in O\}$.
Here $1 \in R^D$ is a vector of all 1. 

Nevertheless, this definition of Pareto regret is with respect to $O$ which requires the reward mean vectors to be constant over time that only holds for stochastic MO-MAB. To this end, we propose a Pareto regret $R_T^{\prime}$ and a Pareto pseudo regret $\bar{R}_T^{\prime}$ based on a similar distance metric but with the newly defined Pareto optimal fronts. To our best knowledge, this is the first work on defining and establishing Pareto regret in adversarial MO-MAB with the generalizability to stochastic MO-MAB as well.

\vspace{-3mm}
\paragraph{Pareto Regret} 
Formally, the Pareto regret in MO-MAB is defined as $R^{\prime}_T = Dist(\sum_{t=1}^Tr^{a_t,t}, O^{\prime})$.
\vspace{-5mm}
\paragraph{Pareto Pseudo Regret} 

In the context of MO-MAB, we propose a Pareto Pseudo regret as $\bar{R}_T^{\prime} = Dist(E[\sum_{t=1}^{T}r^{a_t,t}] , \bar{O}^{\prime})$.

In the proofs and some statements we utilize rewards only along a single dimension. To this end, denote $R_T^d$ as the regret with respect to rewards of dimension or coordinate $d$ as $R_T^d = \max_{i}\sum_{t=1}^{T}r_d^{i,t} - \sum_{t=1}^{T}r_d^{a_t,t}$,
and the corresponding pseudo regret is defined as $\bar{R}_T^d = \max_{i}E[\sum_{t=1}^{T}r_d^{i,t}] - E[\sum_{t=1}^{T}r_d^{a_t,t}]$.

Note that the proposed regrets apply for both stochastic MO-MAB and adversarial MO-MAB since rewards can be arbitrary. To this end, we call $R_T^{\prime}$ the general Pareto regret, different from $R_T$, namely the stochastic Pareto regret that only works for the stochastic setting.

\section{Upper bounds on Pareto regret}\label{sec:ub}

In this section, we formally elaborate on the newly proposed algorithms for MO-MAB based on MAB algorithms, which we consider both with and without a priori knowledge of stochastic or adversarial, and prove their optimality or near optimality in stochastic and adversarial settings simultaneously. More specifically, we denote an indicator for the setting of MO-MAB by $s$ with $s=0$ being stochastic and $s=1$ being 
adversarial. When $s$ is known, we propose Algorithm 1 that achieves optimality. For unknown $s$, i.e. with less information, Algorithm 2 is nearly optimal with respect to Pareto pseudo regret, up to a $\log T$ factor, in line with the work on MAB. 

\subsection{A Priori Knowledge of Stochastic or Adversarial}

When the indicator $s$ is given, Pareto regret in MO-MAB can be related to regret in MAB as follows. 
\begin{theorem}\label{th:regret_d}
 For any dimension $d^{\prime}$, we have that $R_T^{\prime} \leq R_T^{d^{\prime}}$.
\end{theorem}
Based on the result which essentially fully characterizes Pareto regret in the form of vanilla regret in MAB, we formally develop Algorithm~\ref{alg:1} called Multi-Objective with Known S (MO-KS) for MO-MAB, which has comparable performance with MAB by switching between EXP3.P and UCB depending on the problem setting $s$ and a randomly sampled dimension $d^{\prime}$. Theorem~\ref{th:regret_pareto_known} shows the Pareto regret upper bound of Algorithm~\ref{alg:1} with respect to Pareto regret $R^{\prime}_T$, which has the same order as the regret upper bound in MAB. 
\begin{algorithm}[h]

\caption{Algorithm With Known $s$ (MO-KS)}\label{alg:1}
\begin{algorithmic}
\STATE Input: Fixed dimension $d^{\prime}$, $1 \leq d^{\prime} \leq D$;
\STATE Initialization: indicator $s = \{0,1\}$;
\IF{$s$= 0}
     \FOR{$t=1,2,\ldots,T$}
     \STATE Play the bandit game by applying the UCB algorithm along dimension $d^{\prime}$
     \ENDFOR
\ELSIF{$s$= 1}
    \FOR{$t=1,2,\ldots,T$}
     \STATE Play the bandit game by applying EXP3.P algorithm along dimension $d^{\prime}$
     \ENDFOR
 \ENDIF
 \end{algorithmic}
\end{algorithm}
\vspace{-3mm}
\begin{theorem}\label{th:regret_pareto_known}
Based on Algorithm~\ref{alg:1}, for dimension $d^{\prime}$ we have $E[R_T^{d^{\prime}}] \leq O^*(\log{T}) \cdot I_{s =0} + O^*(\sqrt{T}) \cdot I_{s=1}$,
 which leads to $E[R_T^{\prime}] \leq O^*(\log{T}) \cdot I_{s =0} + O^*(\sqrt{T}) \cdot I_{s=1}$.
\end{theorem}

The proof of Theorem~\ref{th:regret_pareto_known} is by combining the result of Theorem~\ref{th:regret_d} and the existing regret upper bounds of the UCB and EXP3.P algorithms.

Note that the choice of $d^{\prime}$ in Algorithm 1 is arbitrary and thus it can depend on the context, since the theoretical guarantee holds for any $d^{\prime}$. For example, for a recommendation system, the click rate may be of more interest for decision makers and thereby being the optimization objective. Moreover, from the perspective of minimizing the constant term in the Pareto regret, we can always specify such a dimension accordingly by running a burning period to determine the optimal dimension, despite of the fact that the regret order remains the same which is a focus of this paper.

\subsection{Lack of Knowledge of Stochastic or Adversarial}

In this section, we focus on MO-MAB where the indicator $s$ is unknown and propose an algorithm (see Algorithm~\ref{alg:2}) that remains nearly optimal, up to a $\log T$ factor, no matter what is the value of $s$. Moreover, Algorithm~\ref{alg:2} allows unknown $T$, compared to the scenarios with known indicators, which increases generalizability.

Similarly to Theorem 1 we have the following result that makes it a possibility to analyze Pareto pseudo regret by means of pseudo regret in the context of MO-MAB. 

\begin{theorem}\label{th:pseudo_regret_d}
For any dimension $d^{\prime}$, we have that $\bar{R}_T^{\prime} \leq \bar{R}_T^{d^{\prime}}.$
\end{theorem}

\begin{algorithm}[h]

\caption{Algorithm With Unknown $s$ and $T$ (MO-US)}\label{alg:2}
\begin{algorithmic}
\STATE Input: Fixed dimension $d^{\prime}$, $1 \leq d^{\prime} \leq D$;
\STATE Initialization: $c =256,\alpha = 3 $;
\STATE Pull each arm once and set $\Tilde{L}_a(K) = r^{a}_{d^{\prime}}$ and $N_a(K) = 1$ for any arm $a$;
\FOR{$t=K+1,K+2,\ldots,T$}
    \STATE $\eta_t = \frac{1}{2}\sqrt{\frac{\ln{K}}{tK}}$;
    \STATE For any arm $a$ let \\
    \quad $UCB_a(t) = \min{\{1,\frac{\Tilde{L}_a(t-1)}{N_a(t-1)} + \sqrt{\frac{\alpha\ln{tK^{\frac{1}{\alpha}}}}{2N_a(t-1)}}\}}$; \\
    \quad $LCB_a(t) = \min{\{1,\frac{\Tilde{L}_a(t-1)}{N_a(t-1)} - \sqrt{\frac{\alpha\ln{tK^{\frac{1}{\alpha}}}}{2N_a(t-1)}}\}}$; \\
    \quad $\zeta_t(a) = \min{\{0,LCB_a(t) - \min_{a^{\prime}}UCB_{a^{\prime}}(t)\}}$; \\
    \quad $\psi_t(a) = \frac{c\ln{t}}{t\zeta_t(a)^2}$; $\epsilon_t(a) = \min{\{\frac{1}{2K},\eta_t,\psi_t(a)\}}$; \\
    \quad $\rho_t(a) = \frac{\exp{-(\eta_t\Tilde{L}_{t-1}(a))}}{\sum_{a^{\prime}}\exp{-(\eta_t\Tilde{L}_{t-1}(a^{\prime}))}}$; \\
    \quad $\Tilde{\rho}_t(a) = (1 - \sum_{a^{\prime}}\epsilon_t(a^{\prime}))\rho_t(a) + \epsilon_t(a) $;
    \STATE Pull an arm  $a_t$ based on probability $\Tilde{\rho}_t(a)$ and receive the reward $r^{a_t,t}$;
    \STATE For any arm $a$ let \\
    \quad $\Tilde{l}_t^a = \frac{1- r^{a_t,t}_{d^{\prime}}}{\Tilde{\rho}_t(a)} \cdot 1_{a = a_t}$;$\Tilde{L}_t(a) =  \Tilde{L}_{t-1}(a) + \Tilde{l}_t^a$; \\
    \quad $N_a(t) = N_a(t-1) + 1_{a = a_t}$;
 \ENDFOR
\end{algorithmic}
\end{algorithm}
The proof of Theorem~\ref{th:pseudo_regret_d} follows the same steps as the proof of Theorem~\ref{th:regret_d} and substitutes the analysis on Pareto regret with the one on Pareto pseudo regret. 

Motivated by the relationship between Pareto pseudo regret and pseudo regret as in Theorem~\ref{th:pseudo_regret_d}, we develop Algorithm~\ref{alg:2} (MO-US) by modifying the EXP3++ algorithm in MAB as in~\citep{seldin2014one, seldin2017improved}. We are given a dimension and apply the parameter-specific EXP3++ algorithm to guarantee near optimality for pseudo regret in that dimension and consequently for Pareto pseudo regret.

The upper bound on Pareto pseudo regret of Algorithm 2 is formally stated as follows.

\begin{theorem}\label{th:pareto_pseudo_regret_unknown}
In Algorithm 2, without knowing the time horizon $T$, for dimension $d^{\prime}$ we have $\bar{R}_T^{d^{\prime}} \leq O^*((\log{T})^2) \cdot I_{s =0} + O^*(\sqrt{T}) \cdot I_{s=1}$.
Pareto Pseudo regret $\bar{R}_T^{\prime}$ can be bounded as $\bar{R}_T^{\prime} \leq O^*((\log{T})^2) \cdot I_{s =0} + O^*(\sqrt{T}) \cdot I_{s=1}$.
\end{theorem}

In like manner, the proof utilizes the result of Theorem~\ref{th:pseudo_regret_d} and the known results of IEXP3++ algorithm as in~\citep{seldin2017improved}. 

Similarly, the discussion aforementioned in the previous subsection on the choice of dimension applies here.

\begin{remark}
Note that there is an algorithm proposed by ~\citep{zimmert2019optimal} to improve the regret order $(\log{T})^2$ in stochastic MAB, which has order $\log{T}$, exactly the same order as the lower bound. However, the additional assumptions of the algorithm  may limit its application in MO-MAB, such as the condition of a unique best arm. For MO-MAB, it is possible that multiple best arms exist in one dimension.
\end{remark}

For stochastic MO-MAB, \citep{drugan2013designing, drugan2014pareto} provide a formulation of regret and subsequently propose optimal algorithms, however their algorithms and analyses do not hold for adversarial settings. While techniques in multi-objective optimization (\citep{hong2021evolutionary,yahyaa2014annealing,busa2017multi,groetzner2022multiobjective}) adapt to any reward distributions, they either convert the MO-MAB problem to single objective MAB by scalarization and minimax or only work empirically. Our MO-MAB results establish algorithms and regret bounds for the adversarial setting, Figure~\ref{fig:gaps}.

\section{Lower bounds on Pareto Regret}\label{sec:lb}

\subsection{Lower Bounds}

In this section, we show that for stochastic MO-MAB and adversarial MO-MAB, there exist scenarios where the regret of any randomized algorithms is exceeding certain lower bounds, with respect to both Pareto regret and Pareto pseudo regret. It validates the optimality of Algorithm~\ref{alg:1} and the near optimality of Algorithm~\ref{alg:2}, up to constant factors. Moreover, the lower bounds stay the same with those for MAB, which further connects MO-MAB with MAB, in conjunction with the upper bound results.  

Formally, the lower bounds on Pareto regret and Pareto pseudo regret in both stochastic MO-MAB and adversarial MO-MAB are summarized as follows.

\begin{theorem}\label{th:pseudo_sto}
For stochastic MO-MAB, there exists scenarios where the Pareto pseudo regret of any randomized algorithms is larger than $\log{T}$. 
\end{theorem}

\begin{theorem}\label{th:pareto_sto}
For stochastic MO-MAB and small $T$, there exist scenarios where the expected Pareto regret of any randomized algorithms is larger than $\log{T}$. 
\end{theorem}

\begin{theorem}\label{th:pseudo_adv}
For adversarial MO-MAB, there exist scenarios where the Pareto pseudo regret of any randomized algorithms is larger than $\sqrt{T}$. 
\end{theorem}

\begin{theorem}\label{th:pareto_adv}
Consider adversarial MO-MAB with repeating values for different dimensions in reward vectors, i.e. $r^{i,t} = (r_1^{i,t}, \ldots, r_1^{i,t})$. Then there exists scenarios where the Pareto regret of any randomized algorithms is larger than $\sqrt{T}$ with high probability. 
\end{theorem}

In the proofs of Theorems ~\ref{th:pseudo_adv} and ~\ref{th:pareto_adv}, we find instances where the regret in MO-MAB is equivalent to the regret in MAB in one dimension, namely the marginal regret. This can be done by letting either the reward vectors or the reward mean vectors have the same numbers for all dimensions. Then the lower bounds on marginal regret essentially hold for MO-MAB when the reward numbers in vectors meet the corresponding conditions in MAB. This argument does not hold for Theorems ~\ref{th:pseudo_sto} and ~\ref{th:pareto_sto} by noting that the reward vectors do not have the same numbers for all dimensions out of stochasticity. This brings additional difficulties and thereby necessitates a new analytical approach. We herein utilize the multi-dimensional concentration inequalities to deal with stochasticity.

For stochastic MO-MAB, \citep{drugan2013designing} establish a lower bound of order $\log T$ and our results are consistent with it, with respect to the newly defined Pareto regret and Pareto pseudo regret. This justifies the introduction of the proposed measures. Furthermore, for adversarial MO-MAB, our regrets have lower bounds of order $\sqrt{T}$, which are the same as in single dimensional MAB~\citep{gerchinovitz2016refined,bubeck2012regret}. Meanwhile, the upper bounds of MO-KS have the same order as the lower bounds, implying optimality of MO-KS, while MO-US is nearly optimal up to a $\log T$ factor since the upper bound in the stochastic setting is $(\log T)^2$.

\subsection{Lower bounds of Pareto UCB}

We have already shown the lower bounds for general algorithms. Motivated by the online adversarial attack on UCB with small attack cost in an MAB setting as studied in~\citep{jun2018adversarial}, for Pareto UCB designed for stochastic MO-MAB, we go further and show that it can lead to linear regrets in terms of Pareto regret $R_T$ and $R_T^{\prime}$ given adaptive adversaries by borrowing the adversarial attack mechanism in MAB. This analysis is made possible by our newly proposed framework of adversarial MO-MAB, highlighting the ineffectiveness of Pareto UCB in such scenarios. Specifically, we consider scenarios where the reward vectors of all arms are sub-Gaussian distributed with the same variance denoted by $\sigma^2$ and $\mu^{K}$ is dominated by any other arm and thus for every 
$i < K,
0 < \Delta_i = \max_{d}\{\mu^i_d - \mu^K_d\}  = ||\mu^i-\mu^K||_{\infty}$. Generally speaking, the adversary, namely Alice, aims
to manipulate Bob, the player, into pulling arm $K$ very often while making small attacks under assumptions that 1) Bob does not know the presence of Alice, 2) the number of arms $K$ and the time horizon $T$ are known to Alice and Bob, 3) Alice knows that arm $K$ is dominated by all other arms, and 4) Alice knows the exact algorithm Bob is using. Note that Alice does not necessarily know the specific arm $a_t$ pulled by Bob at time step $t$, which generalizes the adversarial attack in MAB as in~\citep{jun2018adversarial} that assume a known $a_t$.

Formally, value $\alpha_t$ is the cost of Alice to attack Bob at time step $t$. Parameter $\Delta_0$ is determined by Alice and $\delta \in (0,1)$. (We have already defined $\Delta_i, i \geq 1$ but not $\Delta_0$.) Let $\hat{\mu}^i(t)$ be an estimator for $\mu^i$ up to time step $t$, while $\hat{\Tilde{\mu}}^i(t)$, $\Tilde{\mu}^i = \mu^i - \bar{\alpha}_t = \mu^i - \dfrac{\sum_{t = 1}^T{\alpha_t}}{T}$ are the estimator and true value of the post-attack reward of arm $i$, respectively. 
\begin{algorithm}[h]
\caption{Attack UCB}
\label{mab:attack}
\begin{algorithmic}
\STATE Initialization: $\beta(n) = \sqrt{\frac{2\sigma^2}{n}\log{\frac{\pi^2Kn^2}{3\delta}}}$;
 \FOR{$t=1,2,\ldots,T$}
     \STATE Bob chooses $a_t$ according to the UCB algorithm;
     \STATE Simultaneously Alice computes $a_t$ based on past $t-1$ observations and the UCB algorithm;
     \STATE The environment generates rewards for each arm $i$ by sampling from a stochastic generator;
     \STATE Alice learns the pre-attack reward $r^{a_t,t}$ from the environment;
     \STATE By using $r^{a_t,t}$ Alice updates all $\hat{\mu},\Tilde{\hat{\mu}}$;
     \STATE $\alpha_t = 1_{a_t = K} \cdot \max \{0,\hat{\mu}^K(t) - 2\beta(N_{K}(t))$ \\ \qquad \quad $- \Delta_0 - \hat{\Tilde{\mu}}^{a_t}(t)\}$;
   \STATE Bob receives reward $r^{a_t,t} - \alpha_t$;
   \ENDFOR
\end{algorithmic}
\end{algorithm}
The attack algorithm in MAB is shown in Algorithm~\ref{mab:attack}. At each time step, Alice predicts the choice $a_t$ of Bob given by the UCB algorithm, and attacks $a_t \neq K$ to mislead Bob. 

When it comes to Pareto UCB in the context of MO-MAB, randomness in the Pareto optimal front presents a concern since $a_t$ is unpredictable. As a generalization, we propose Algorithm~\ref{mo-mab:attack} that attacks Pareto UCB with small costs. Let $O^{\prime}, \bar{O}_t^{\prime}$ be the post-attack Pareto optimal front, let $O$ again be the pre-attack Pareto optimal front on $\mu^{i}$ and $O_t$ be the estimator for $O$ up to time step $t$. As before we similarly denote the quantities with subscript $A$ to capture arm indices. More specifically, we modify the estimation for the best arm in UCB by estimating Pareto optimal front $O^t $ and attack any arm in $O_A^t$ if $K \not\in O_A^t$ and determine the attack cost as a corruption by the Pareto order relationship. 

\subsubsection{Results on Stochastic Pareto Regret}

Under Algorithm~\ref{mo-mab:attack}, we present a lower bound on stochastic Pareto Regret $R_T$ defined in ~\cite{drugan2013designing}. We assume that Alice does not attack in the first $2K$ rounds, i.e. $\sum_{s=1}^{2K}\alpha_s = 0$. The result generalizes the adversarial attack on UCB to adversarial attack on Pareto-UCB, i.e. from MAB to MO-MAB. We first state Theorem~\ref{th:non-target} which provides an upper bound on the number of times Bob does not pull arm $K$ and on the total attack cost. 
\begin{algorithm}[h]
\caption{Attack Pareto UCB}
\label{mo-mab:attack}
\begin{algorithmic}
\STATE Initialization: $\beta(n) = \sqrt{\frac{2\sigma^2}{n}\log{\frac{\pi^2Kn^2}{3\delta}}}$;
 \FOR{$t=1,2,\ldots,T$}
     \STATE Bob computes the Pareto front $O_t$;
     \STATE The environment generates reward vector for each arm by sampling from a generator;
     \STATE Simultaneously Alice computes the Pareto front $O_t$ based on past $t-1$ observations and the Pareto UCB algorithm;
     \STATE Alice learns the pre-attack reward $r^{i,t}$ for $i \in O_A^t$;
     \STATE By using $r^{i,t}, i \in O_A^t$, Alice updates $ \hat{\mu},  \hat{\Tilde{\mu}}$;
     \IF{$K \in O^t_A$}
     \STATE $\alpha_t = 0$ \;
     \ELSE 
       \FOR{$j \in O_A^t$}
       \STATE $\bar{z}^j =\hat{\mu}^K - (2\beta(N_{K}(t)) + \Delta_0)\cdot 1 $;
       \STATE $\hat{z}^j = \frac{N_j(t-1)(\hat{\mu}^j(t-1)) - \sum_{s=1}^{t-1}\alpha_s \cdot 1 + r^{j,t}}{N_j(t)}$;
       \ENDFOR
     \ENDIF
    \STATE $\alpha_t =\max\{   \max_{j \in O_A^t, d} \{N_j(t)(\hat{z}^j_d - \bar{z}^j_d)\},0\}$;
   \STATE Bob randomly samples arm $a_t$ from $O_A^t$ and receives reward $r^{a_t,t} - \alpha_t \cdot 1$;
   \ENDFOR
\end{algorithmic}
\end{algorithm}
\vspace{-3mm}
\begin{theorem} \label{th:non-target}
Let $K \geq \frac{3e^2 \delta }{\pi^2}$. With probability $1-D\delta$, for any $T > 2K$, under Algorithm~\ref{mo-mab:attack}, any non-target arm is pulled $O^*(\log{T})$ times and the total attack cost is $(K-1)\left(2 + \frac{9\sigma^2}{\Delta_0^2}\log{T}\right) \cdot \max_i (\Delta_i + \Delta_0) + O^*(\log{T}) $.
\end{theorem}

As a consequence, the Pareto regret $R_T$ can be bounded by its definition, which is stated as Theorem~\ref{th:Pareto-regret}. 

\begin{theorem} \label{th:Pareto-regret}
With probability $1-D\delta$, for any $T > 2K$ and $K \geq \frac{3e^2\delta}{\pi^2}$, the Pareto regret $R_T$ of the Pareto UCB algorithm is at least of order $O^*(T)$ under the adversarial attack generated by Algorithm~\ref{mo-mab:attack}. 
\end{theorem}

The proofs of Theorem~\ref{th:non-target} and Theorem~\ref{th:Pareto-regret} follow the logic of what have been established in~\citep{jun2018adversarial}, but the analyses are mostly in the reward vector space and under our newly proposed attack algorithm. The dimensionality of rewards presents non-trivial challenges in that the attack is on multiple arms at each time step and Bob's rule of playing is Pareto UCB without deterministic choices. To this end, we use multi-dimensional concentration inequalities, \citep{drugan2013designing}. 

For stochastic MO-MAB under no adversarial attack, Pareto UCB is optimal with respect to the Pareto regret $R_T$ as established in~\citep{drugan2013designing}, the optimality of which is unknown in adversarial settings. While in~\citep{jun2018adversarial}, the adversarial attack on UCB in MAB is fully studied, it does not apply to Pareto UCB for MO-MAB. When adversarial attack is added to Pareto UCB, which leads to an adversarial MO-MAB setting, we show that its Pareto regret $R_T$ is at least linear in $T$.

\subsubsection{Results on General Pareto Regret}

Next, we study the lower bound on the general Pareto regret $R^{\prime}_T$ for Pareto UCB under Algorithm~\ref{mo-mab:attack}, as a validation for the claim that Pareto UCB is not optimal for adversarial MO-MAB.

Let us denote quantity $\bar{\alpha}_t^j$ to be the counter-factual attack cost of arm $j$ with respect to arm $K$. They are computed by Alice only for the arms in the Pareto optimal set and are 0 for the arms not in the Pareto optimal set. In other words, the counter-factual attack cost $\bar{\alpha}_t^j$ is defined for each arm $j$ as $\bar{\alpha}_t^i =  \max \{ \max_{d} N_i(t)(\hat{z}^i_d - \bar{z}^i_d),0\}$. Based on Algorithm~\ref{mo-mab:attack}, $\alpha_t = \max_{j \in O_A^t}\bar{\alpha}_t^j = \max_{j}\bar{\alpha}_t^j$.  


We start by properly defining the Pareto optimal front for stochastic MO-MAB with the attack cost, which is necessary for Pareto regret $R^{\prime}_T$. Two options can be considered: one averaging the actual attack costs and the second taking into account the counterfactual attack costs.  
\vspace{-4mm}
\paragraph{Definition 1}
Given adversarial attack cost, $\bar{O}^{\prime}$ is defined as the Pareto optimal front of vectors $\mu^{i} - \frac{1}{\sum_{j \neq K}N_j(T)} \cdot \sum_{a_t \neq K}\alpha_t \cdot 1$ over all arm $i$, whereas $O^{\prime}$ is over vectors $\frac{1}{T}\sum_{t=1}^T r^{i,t} - \frac{1}{N_i(T)}\sum_{i_t = i}\alpha_t \cdot 1$. 
\vspace{-5mm}
\paragraph{Definition 2}
Given adversarial attack cost, $\bar{O}^{\prime}$ is defined as the Pareto front of vectors 
$\mu^i - \frac{1}{T} \sum_{t=1}^T\bar{\alpha}^i_t \cdot 1$, whereas $O^{\prime}$ is over $\frac{1}{T}(\sum_{t=1}^T r^{i,t} - \sum_{t =1}^T\bar{\alpha}^i_t \cdot 1)$. 
\vspace{-3mm}
\paragraph{Assumption 4}
Let $S$ be the set of $\{\mu^1,\ldots,\mu^K\}$. We assume that the distance from the arms that are not in Pareto front of $S$ to the arms in Pareto front of $S$ is at least $5\gamma$ for $0 < \gamma < \frac{1}{5}$. Formally, $\mu^j \succeq \mu^i + 5\gamma \cdot 1$
holds for any arm $i$ not in Pareto front of $S$ and arm $j$ in Pareto front of $S$.

Based on either definition, we establish a lower bound on Pareto regret under certain conditions as follows. First, we consider the high probability regret lower bound which essentially implies a lower bound on the expected regret by integration.

\begin{theorem} \label{th:Pareto-regret-new-1}
Under Definition 1 and Assumption 4, with probability $1-2\eta-D\delta$, for any $T$, $0 < \mu < 1$, $\delta \leq \frac{2\pi^2}{3e^2}$, and $\sigma^2 \leq \frac{1}{5(\ln{(4D)} + \ln{\frac{1}{\mu}})}$, the Pareto regret $R^{\prime}_T$ of the Pareto UCB algorithm is at least of order $O(T)$ based on Algorithm~\ref{mo-mab:attack}, if $K=2$ and $\gamma \leq \sqrt{2\sigma^2(\ln{(4D)} + \ln{\frac{1}{\mu}})}$. 
\end{theorem}

The proof connects $R_T^{\prime}$ and $R_T$ by concentration inequalities and an asymptotic property of the attack cost and then studying $R_T$. Next based on Definition 2, we obtain a similar result.

\begin{theorem} \label{th:Pareto-regret-new-2}
Under Definition 2 and Assumption 4, with probability $(1-2\eta-D\delta)$, for $T \geq T(\gamma)$, $0 < \mu < 1$, $\delta \leq \frac{2\pi^2}{3e^2}$ and $\sigma^2 \leq \frac{1}{5(\ln{(4D)} + \ln{\frac{1}{\mu}})}$, the Pareto regret $R^{\prime}_T$ of the Pareto UCB algorithm is at least of order $O(T)$ based on Algorithm~\ref{mo-mab:attack} if $K=2$ and $\gamma \leq \sqrt{2\sigma^2(\ln{(4D)} + \ln{\frac{1}{\mu}})}$.
\end{theorem}

Again, for the proof we utilize concentration inequalities and the asymptotic property of attack cost and the lower bound on $R_T$ aforementioned.

For Pareto regret in~\citep{drugan2013designing}, we previously establish that its lower bound is linear when attacking Pareto UCB with adversarial costs by extending the MAB result in~\citep{jun2018adversarial}. The regret analysis~\citep{jun2018adversarial} only covers pre-attack rewards, while the post-attack scenarios are not studied. Here we generalize the result to our newly defined Pareto regret and Pareto pseudo regret defined on post-attack reward vectors, to further validate non-optimality of Pareto UCB in adversarial MO-MAB.

\begin{remark}
It is worth highlighting the robustness of our newly proposed algorithm against this online adversarial attack. In the presence of such attacks, where the attack cost is of order $\log T$ as shown in Theorem~\ref{th:non-target} and supported by Lemma~\ref{le:N}, we observe that for any large $t$, $N_i(t) \leq \min{\{N_K(t), 2 + \frac{9\sigma^2}{\Delta_0^2}\log{t}}\}$. This implies that $N_K(t) = O(T)$. As a result, for at least $O(T)$ rounds, Alice refrains from launching attacks. During this period, the regret of our proposed algorithms is of order $O(\sqrt{T})$, which is different from Pareto UCB that exhibits linear regret in such scenarios. 
\end{remark}
\section{Conclusion}

\begin{figure}[h]
    \centering
\includegraphics[width=0.47\textwidth]{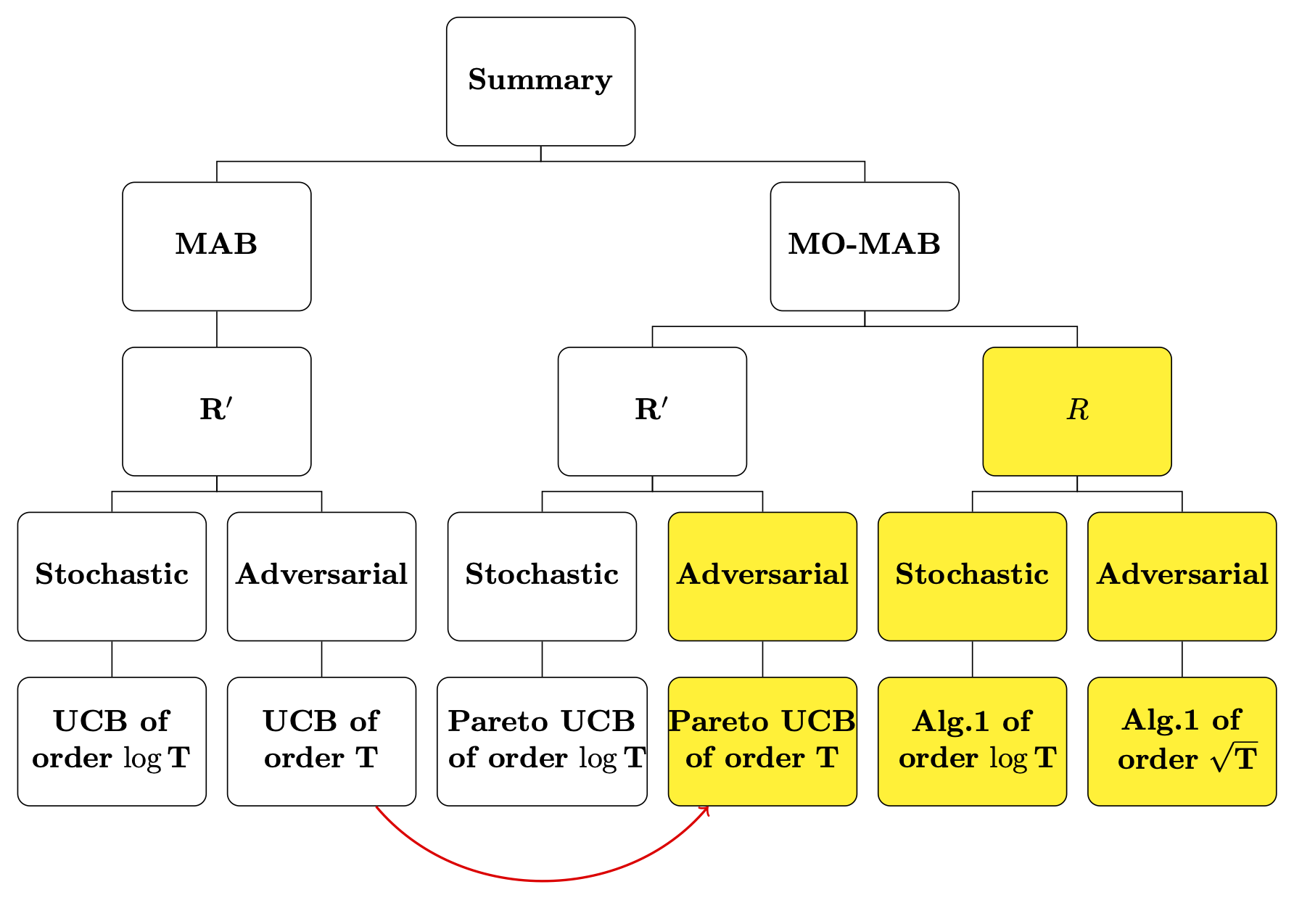}
    \caption{An illustration of our contributions}
    \label{fig:gaps}
\end{figure}

In this paper, we study multi-objective multi-armed bandit (MO-MAB) with full Pareto regret analyses. We formulate the general MO-MAB problem for both stochastic and adversarial settings from a perspective of Pareto optimality and regret. Then we propose an optimal algorithm when given a priori knowledge of stochastic and adversarial and develop a nearly optimal algorithm up to a $\log T$ factor when lack of such knowledge is present. The theoretical guarantee is fully examined by both upper bounds and lower bounds on Pareto regret and Pareto pseudo regret. Moreover, we extend the adversarial attack algorithm for UCB to a new one attacking Pareto UCB in the context of MO-MAB. We also establish a lower bound on stochastic Pareto regret for Pareto UCB and subsequently show that Pareto UCB may have poor performance in adversarial settings by providing a lower bound on general Pareto regret. A summary of the research gaps we close is shown in Figure~\ref{fig:gaps} where the highlighted parts are the contributions of this paper.

\begin{table}[H]
  \caption{Comparisons}\label{tb:1}
  \centering 
  \begin{threeparttable}
    \begin{tabular}{ccc}
    Area  & Application  & Results\\
    \midrule\midrule
    (1)  &  \makecell{Pareto order relationship;\\ Pareto optimal front;\\ MO-MAB} & $\sqrt{T\ln{T}}$ \\
    \cmidrule(l  r ){1-3}
    (2) & \makecell{Attacks on \\ stochastic \\ bandits} & \makecell{regret \\ $T$} \\
    \cmidrule(l  r ){1-3}
    (3) & \makecell{(nearly) optimal \\ in  the (stochastic) \\ adversarial settings} & \makecell{$\ln{T}$ ($\ln^2{T}$);\\$\sqrt{T}$}
    \end{tabular}
    \begin{tabular}{ccc}
      \midrule\midrule
    Area & Limitation & Our results \\
    \midrule\midrule
    (1) & \makecell{scalarization functions; \\ solving $T$  subproblems} & \makecell{formulation of \\general \\MO-MAB} \\
    \cmidrule(l  r ){1-3}
    (2) & \makecell{single-objective; \\ regret not in \\ adversarial settings} & \makecell{analyses in \\MO-MAB; \\ regret $T$} \\
    \cmidrule(l  r ){1-3}
    (3) & single-objective & \makecell{ analyses  in \\ MO-MAB; \\$\ln{T}$ ($\ln^2{T}$);\\$\sqrt{T}$} \\
    \midrule\midrule
    \end{tabular}
\end{threeparttable}
  \end{table}

To explore and compare the connections between the research domains discussed herein, we present a comprehensive summary table (Table \ref{tb:1}) that examines three distinct areas of research: Pareto optimization (labeled as (1)), adversarial attack (labeled as (2)), and best-of-both-worlds algorithms (labeled as (3)). We compare these methods from various perspectives, including their applications, known results, and areas where they have not yet been studied, or equivalently, potential limitations, as well as our own contributions to these fields. For simplicity, we use the term 'nearly optimal' to denote performance up to a factor of $\log{T}$. 

As a future work, we point out that it could be of great interest to improve the upper bound of Pareto pseudo regret to achieve optimality given no knowledge of MO-MAB settings. Furthermore,  this is in line with the current work on MAB, since it remains open to develop an optimal algorithm for general MAB under no assumptions. Meanwhile, it is worth investigating the explicit relationship between the regret upper bound and both the dimension $D$ and the size of the Pareto optimal set $|O|$. The existing lower bound results already indicate a dependency on $D$, suggesting the possibility of enhancing the regret upper bound through the development of more effective algorithms.  As a concluding remark, other metrics can be developed for measuring the performance of algorithms, such as unfairness. 
\section{Acknowledgement}

We would like to thank the ICML anonymous reviewers and meta-reviewer for their helpful suggestions and valuable comments. Their feedback has greatly helped to improve the paper.

\bibliography{icml_2023}

\begin{thebibliography}{20}
\providecommand{\natexlab}[1]{#1}
\providecommand{\url}[1]{\texttt{#1}}
\expandafter\ifx\csname urlstyle\endcsname\relax
  \providecommand{\doi}[1]{doi: #1}\else
  \providecommand{\doi}{doi: \begingroup \urlstyle{rm}\Url}\fi

\bibitem[Audibert and Bubeck(2009)]{audibert2009minimax}
J.-Y. Audibert and S.~Bubeck.
\newblock Minimax policies for adversarial and stochastic bandits.
\newblock In \emph{Conference on Learning Theory}, volume~7, pages 1--122,
  2009.

\bibitem[Auer and Chiang(2016)]{auer2016algorithm}
P.~Auer and C.-K. Chiang.
\newblock An algorithm with nearly optimal pseudo-regret for both stochastic
  and adversarial bandits.
\newblock In \emph{Conference on Learning Theory}, pages 116--120. Proceedings
  of Machine Learning Research, 2016.

\bibitem[Auer et~al.(2002)Auer, Cesa-Bianchi, Freund, and
  Schapire]{auer2002nonstochastic}
P.~Auer, N.~Cesa-Bianchi, Y.~Freund, and R.~E. Schapire.
\newblock The nonstochastic multiarmed bandit problem.
\newblock \emph{SIAM Journal on Computing}, 32\penalty0 (1):\penalty0 48--77,
  2002.

\bibitem[Bubeck et~al.(2012)Bubeck, Cesa-Bianchi, et~al.]{bubeck2012regret}
S.~Bubeck, N.~Cesa-Bianchi, et~al.
\newblock Regret analysis of stochastic and nonstochastic multi-armed bandit
  problems.
\newblock \emph{Foundations and Trends{\textregistered} in Machine Learning},
  5\penalty0 (1):\penalty0 1--122, 2012.

\bibitem[Busa-Fekete et~al.(2017)Busa-Fekete, Sz{\"o}r{\'e}nyi, Weng, and
  Mannor]{busa2017multi}
R.~Busa-Fekete, B.~Sz{\"o}r{\'e}nyi, P.~Weng, and S.~Mannor.
\newblock Multi-objective bandits: Optimizing the generalized {G}ini index.
\newblock In \emph{International Conference on Machine Learning}, pages
  625--634. Proceedings of Machine Learning Research, 2017.

\bibitem[Drugan and Nowe(2013)]{drugan2013designing}
M.~M. Drugan and A.~Nowe.
\newblock Designing multi-objective multi-armed bandits algorithms: A study.
\newblock In \emph{The 2013 International Joint Conference on Neural Networks},
  pages 1--8. IEEE, 2013.

\bibitem[Drugan et~al.(2014)Drugan, Now{\'e}, and Manderick]{drugan2014pareto}
M.~M. Drugan, A.~Now{\'e}, and B.~Manderick.
\newblock Pareto upper confidence bounds algorithms: an empirical study.
\newblock In \emph{2014 IEEE Symposium on Adaptive Dynamic Programming and
  Reinforcement Learning}, pages 1--8. IEEE, 2014.

\bibitem[Gerchinovitz and Lattimore(2016)]{gerchinovitz2016refined}
S.~Gerchinovitz and T.~Lattimore.
\newblock Refined lower bounds for adversarial bandits.
\newblock \emph{Advances in Neural Information Processing Systems}, 29, 2016.

\bibitem[Groetzner and Werner(2022)]{groetzner2022multiobjective}
P.~Groetzner and R.~Werner.
\newblock Multiobjective optimization under uncertainty: A multiobjective
  robust (relative) regret approach.
\newblock \emph{European Journal of Operational Research}, 296\penalty0
  (1):\penalty0 101--115, 2022.

\bibitem[Hong et~al.(2021)Hong, Yang, and Tang]{hong2021evolutionary}
W.-J. Hong, P.~Yang, and K.~Tang.
\newblock Evolutionary computation for large-scale multi-objective
  optimization: A decade of progresses.
\newblock \emph{International Journal of Automation and Computing}, 18\penalty0
  (2):\penalty0 155--169, 2021.

\bibitem[H{\"u}y{\"u}k and Tekin(2021)]{huyuk2021multi}
A.~H{\"u}y{\"u}k and C.~Tekin.
\newblock Multi-objective multi-armed bandit with lexicographically ordered and
  satisficing objectives.
\newblock \emph{Machine Learning}, 110\penalty0 (6):\penalty0 1233--1266, 2021.

\bibitem[Jun et~al.(2018)Jun, Li, Ma, and Zhu]{jun2018adversarial}
K.-S. Jun, L.~Li, Y.~Ma, and J.~Zhu.
\newblock Adversarial attacks on stochastic bandits.
\newblock \emph{Advances in {N}eural {I}nformation {P}rocessing {S}ystems}, 31,
  2018.

\bibitem[Lattimore and Szepesv{\'a}ri(2020)]{lattimore_szepesvari_2020:ch16}
T.~Lattimore and C.~Szepesv{\'a}ri.
\newblock \emph{Bandit Algorithms}.
\newblock Cambridge University Press, 2020.
\newblock \doi{10.1017/9781108571401}.

\bibitem[Mehrotra et~al.(2020)Mehrotra, Xue, and Lalmas]{mehrotra2020bandit}
R.~Mehrotra, N.~Xue, and M.~Lalmas.
\newblock Bandit based optimization of multiple objectives on a music streaming
  platform.
\newblock In \emph{Proceedings of the 26th ACM SIGKDD {I}nternational
  {C}onference on {K}nowledge {D}iscovery \& {D}ata {M}ining}, pages
  3224--3233, 2020.

\bibitem[Seldin and Lugosi(2017)]{seldin2017improved}
Y.~Seldin and G.~Lugosi.
\newblock An improved parametrization and analysis of the {EXP3}++ algorithm
  for stochastic and adversarial bandits.
\newblock In \emph{Conference on Learning Theory}, pages 1743--1759.
  Proceedings of Machine Learning Research, 2017.

\bibitem[Seldin and Slivkins(2014)]{seldin2014one}
Y.~Seldin and A.~Slivkins.
\newblock One practical algorithm for both stochastic and adversarial bandits.
\newblock In \emph{International Conference on Machine Learning}, pages
  1287--1295. Proceedings of Machine Learning Research, 2014.

\bibitem[Van~Moffaert et~al.(2014)Van~Moffaert, Van~Vaerenbergh, Vrancx, and
  Now{\'e}]{van2014multi}
K.~Van~Moffaert, K.~Van~Vaerenbergh, P.~Vrancx, and A.~Now{\'e}.
\newblock Multi-objective $\chi$-armed bandits.
\newblock In \emph{2014 International Joint Conference on Neural Networks},
  pages 2331--2338. IEEE, 2014.

\bibitem[Yahyaa et~al.(2014)Yahyaa, Drugan, and Manderick]{yahyaa2014annealing}
S.~Q. Yahyaa, M.~M. Drugan, and B.~Manderick.
\newblock Annealing-pareto multi-objective multi-armed bandit algorithm.
\newblock In \emph{2014 IEEE Symposium on Adaptive Dynamic Programming and
  Reinforcement Learning}, pages 1--8. IEEE, 2014.

\bibitem[Zhu and Nowak(2020)]{zhu2020regret}
Y.~Zhu and R.~Nowak.
\newblock On regret with multiple best arms.
\newblock \emph{Advances in Neural Information Processing Systems},
  33:\penalty0 9050--9060, 2020.

\bibitem[Zimmert and Seldin(2019)]{zimmert2019optimal}
J.~Zimmert and Y.~Seldin.
\newblock An optimal algorithm for stochastic and adversarial bandits.
\newblock In \emph{The 22nd International Conference on Artificial Intelligence
  and Statistics}, pages 467--475. Proceedings of Machine Learning Research,
  2019.

\end{thebibliography}

\clearpage
\newpage

\section{Appendix}

\subsection{Proofs of the results in Section~\ref{sec:ub}}

Throughout the following lemmas, we let $a$ be a vector such that the set $\{\epsilon \geq 0: a+ \epsilon 1 || \sigma \text{ for every }\sigma \in O \}$ is not empty. 

\begin{lemma}\label{lemma:1}
For any $a$ and $O$ we have 
\begin{align*}
    u = Dist(a,O) = \min_{d}\max_{\sigma \in O}(\sigma_d - a_d).
\end{align*}
\end{lemma}

\begin{proof}
By definition, we have that
\begin{align*}
    Dist(a,O) & = \min_{\epsilon \geq 0}\{\epsilon: a +\epsilon1 || \sigma \text{ for every } \sigma \in O\}  \\
    & = \min_{\epsilon \geq 0}(A_\epsilon).
\end{align*}

For $\epsilon \in A_\epsilon$ and for every $\sigma \in O$, there exists at least one dimension $d$ such that $a_d + \epsilon > \sigma_d$ and one dimension $d^{\prime}$ such that $a_{d^{\prime}} + \epsilon \leq \sigma_{d^{\prime}}$.

Let $B$ be the set of $\epsilon$ where for every $\sigma \in O$, there exists a dimension $d$ such that $\epsilon \leq \sigma_d - a_d$. Similarly, let $C$ be the set of $\epsilon$ where for every $\sigma \in O$, there exists a dimension $d^{\prime}$ such that $\epsilon > \sigma_{d^{\prime}}- a_{d^{\prime}}$. Clearly, we observe that 
\begin{align*}
    A_\epsilon = B \cap C.
\end{align*}
Formally, we have $B = \{\epsilon: 0 \leq \epsilon \leq b = \min_{\sigma}\max_d(\sigma_d - a_d)\}$ and $C = \{\epsilon: \epsilon \geq c = \min_{d}\max_{\sigma}(\sigma_d - a_d)\}$, by their definitions. Note that for $A_{\epsilon} \neq \emptyset$, it must be the case that $c \leq b$. Therefore, it indicates  
\begin{align*}
   \min_\epsilon(A_\epsilon) & = \min_\epsilon(B \cap C) = c,
\end{align*}
which completes the proof.

\end{proof}

\begin{lemma}~\label{lemma:2}
If $O = \{\sigma\}$, then $Dist(a,O) = \min_d (\sigma_d - a_d)$.
\end{lemma}
\begin{proof}
This is a consequence of letting $O = \{\sigma\}$ in Lemma~\ref{lemma:1}.

\end{proof}

\begin{lemma}\label{lemma:3}
There exists $\bar{\sigma} \in O$ such that $a + u1 \preceq \bar{\sigma}$ where $u$ is defined in Lemma~\ref{lemma:1}. For any dimension $d$ we have 
\begin{align*}
  u \leq \max_{\sigma \in O} (\sigma_d - a_d).
\end{align*}
\end{lemma}

\begin{proof}

We prove the first part by contradiction. Let us assume that $a + u1 || \sigma$ for every $\sigma \in O$. Then for each $\sigma$ there exist $d = d(\sigma)$ such that $a_d + u > \sigma_d$. We denote by $s(\sigma)$ the set of all dimensions with this property. We have $s(\sigma) \neq \emptyset$, $(s(\sigma))^c \neq \emptyset$. Let us define $l(\sigma) = \max_{d \in s(\sigma)}(a_d+u-\sigma_d)$ and $d^{*}(\sigma) = \argmax_{d \in s(\sigma)} (a_d+u-\sigma_d)$. By definition $l(\sigma) >0$. Let $\epsilon = \min_{\sigma \in O}l(\sigma) >0$.

We consider $a + (u-\frac{\epsilon}{2})1$. For every $\sigma \in O$ and $d \in d^{*}(\sigma) \neq \emptyset$, we have $a_d + u - \frac{\epsilon}{2} \geq a_d + u - \frac{l(\sigma)}{2} = a_d + u - \frac{a_d+u - \sigma_d}{2} > \sigma_d$. For $d \not \in s(\sigma)$ (note that $s(\sigma)^c \neq \emptyset$), we have $a_d + u - \frac{\epsilon}{2} \leq a_d + u \leq \sigma_d$.

We conclude that $a+(u-\frac{\epsilon}{2})1 || \sigma$ for every $\sigma \in O$ which contradicts the definition of $u$.

For the second part, $a + u1 \preceq \bar{\sigma} $ implies that for every $d$ we have $u \leq \bar{\sigma}_d - a_d$. The existence of $\bar{\sigma}$ gives us that $u \leq \max_{\sigma \in O}(\sigma_d - a_d)$.

\end{proof}

\subsubsection{Proof of Theorem~\ref{th:regret_d}}
\begin{proof}

It follows from Lemma~\ref{lemma:3} and definitions. Specifically, we use Lemma~\ref{lemma:3} by applying $a = \sum_{t=1}^Tr^{a_t,t}$ and $O = \sum_{t=1}^Tr^{i^*,t}$, where $a$ meets the condition that $a \not\in \{a \succeq \sigma  \text{ for some } \sigma \in O\}$. Otherwise, by the fact that $\sum_{t=1}^Tr^{a_t,t}$ does not dominate any $\sum_{t=1}^Tr^{i^*,t} \in O$ which is the set of Pareto optimal reward vectors, we have $a + c1 = \sigma$ for exactly one $c$ and $\sigma \in O$. In this case, the distance is equivalent to $\sigma_d - a_d \leq \max_{\sigma \in O}(\sigma_d - s_d)$. 

\end{proof}

\subsubsection{Proof of Theorem~\ref{th:regret_pareto_known}}

\begin{proof}
The result can be shown by combining the regret analysis of the EXP3.P and UCB algorithms in one-dimension MAB. 

Formally, when $s=0$, regret $R_T^{d^{\prime}}$ of UCB satisfies that $E[R_T^{d^{\prime}}] \leq O^*(\log{T})$.

While $s=1$ gives us that the regret of EXP3.P satisfies $E[R_T^{d^{\prime}}] \leq O^*(\sqrt{T})$.

Meanwhile, according to Theorem~\ref{th:regret_d}, we have that $R_T \leq \min_d{R_T^d}$. Then by the Jensen's inequality, the expected value of $R_T$ can be upper bounded by 
\begin{align}\label{eq:e_r_t}
    E[R_T] & \leq E[\min_d{R_T^d}] \leq \min_d{E[R_T^d]} \leq E[R_T^{d^{\prime}}].
\end{align}
To conclude, the result follows by plugging the upper bounds on $E[R_T^{d^{\prime}}]$ in~(\ref{eq:e_r_t}).

\end{proof}

\subsubsection{Proof of Theorem~\ref{th:pseudo_regret_d}}

\begin{proof}
It follows from Lemma~\ref{lemma:3} and definitions. Likewise, we again leverage Lemma~\ref{lemma:3} by using $a = \sum_{t=1}^TE[r^{a_t,t}]$ and $O = \sum_{t=1}^TE[r^{i^*,t}]$, where $a$ meets the condition that $a \not\in \{a \succeq \sigma \text{ for some } \sigma \in O\}$. Otherwise, by noting that $\sum_{t=1}^TE[r^{a_t,t}]$ does not dominate any $\sum_{t=1}^TE[r^{i^*,t}] \in O$ which is the set of Pareto optimal expected reward vectors, we obtain  $a + c1 = \sigma$ for exactly one $c$ and $\sigma \in O$ and thus the distance equals to $\sigma_d - a_d \leq \max_{\sigma \in O}(\sigma_d - s_d)$, which completes the proof. 

\end{proof}

\subsubsection{Proof of Theorem~\ref{th:pareto_pseudo_regret_unknown}}

\begin{proof}

By the choice of parameters $\eta_t, \zeta_t(a), \psi_t(a)$, the results hold as follows.
In the adversarial regime, we have that the pseudo regret satisfies
\begin{align*}
    \bar{R}_T^{d^{\prime}} \leq O^*(\sqrt{T}),
\end{align*}
while in the stochastic regime, the pseudo regret satisfies 
\begin{align*}
    \bar{R}_T^{d^{\prime}} \leq O^*((\log{T})^2).
\end{align*}

Meanwhile, we have that 
\begin{align}\label{ineq:r}
    \bar{R}_T^{\prime} & \leq \min_d{\bar{R}_T^{d}} \leq \bar{R}_T^{d^{\prime}},
\end{align}
where the first inequality holds by the argument as in the proof of  Theorem~\ref{th:pseudo_regret_d}. 

By plugging the upper bounds of $\bar{R}_T^{d^{\prime}}$ in~\ref{ineq:r}, we derive the statement in the theorem.

\end{proof}

\subsection{Proofs of the results in Section~\ref{sec:lb}}

\subsubsection{Proof of Theorem~\ref{th:pseudo_sto}}

\begin{proof}

We consider stochastic MO-MAB with reward vectors $\mu^{i} = (\mu_1^{i}, \ldots, \mu_1^{i}) \in R^D$ for any arm $i$ and any time steps $t$. By the definition of $\bar{O}^{\prime}$, we have that $\bar{O}^{\prime}$ is equivalent to a unique best arm $i^*$ with $\mu_1^{i^*} = \max_i\mu_1^{i}$. It gives us that $\bar{R}_T^{\prime} = \bar{R}_T^1 = \ldots = \bar{R}_T^D$ by noting that for any dimension $d$ and by Lemma 2 we have 
\begin{align*}
    \bar{R}_T^{\prime} & =  Dist(E[\sum_{t=1}^{T}r^{a_t,t}] , \bar{O}^{\prime}) \\
    & = Dist(E[\sum_{t=1}^{T}r^{a_t,t}] , \{T\mu^{i^*}\}) \\
    & = Dist(\sum_{t=1}^{T}\mu^{a_t} , \{T\mu^{i^*}\}) \\ 
    & =  T \cdot \mu_d^{i^*} - \sum_{t=1}^{T}\mu_1^{a_t} = \bar{R}_T^d.
\end{align*}

By the result in~\citep{bubeck2012regret} stating that 
\begin{align*}
    \lim\inf_T \frac{\bar{R}_T^1}{\log{T}} \geq O(1),
\end{align*}
we have 
\begin{align*}
    \lim\inf_T \frac{\bar{R}_T^{\prime}}{\log{T}} \geq O(1).
\end{align*}

\end{proof}

\subsubsection{Proof of Theorem~\ref{th:pareto_sto}}

\begin{proof}

We can assume $N_i(T) \geq 1$ since otherwise the underlying algorithm has linear regret. Consider stochastic MO-MAB with reward mean vectors being $\mu^{i} = \mu_1^{i} \cdot 1$ for any arm $i$ and any time steps $t$. It indicates that $r_j^{i,t},r_k^{i,t}$ are i.i.d for any $1 \leq j,k \leq D$, though potentially having different values as a result of stochasticity.  Therefore, the difference between $R_T^{\prime}$ and $R_T^1$ can be bounded by the concentration inequality as follows.

Note that by Lemma~\ref{lemma:2} and ~\ref{lemma:3}, Theorem~\ref{th:regret_d} and Theorem~\ref{th:pseudo_regret_d},we have 
\begin{align} \label{eq:r_t_prime}
    R_T^{\prime} & = Dist(\sum_{t=1}^Tr^{a_t,t},O^{\prime}) \notag \\
    & = \min_d\max_i (\sum_tr^{i,t}_d - \sum_{t=1}^Tr^{a_t,t}_d) \notag \\
    & =  \min_d\max_i(\sum_{t=1}^Tr^{i,t}_d - \sum_{j=1}^K\sum_{a_t=j}r^{j,t}_d)
\end{align}
and 
\begin{align*}
\bar{R}_T^{\prime} & = Dist(E[\sum_{t=1}^{T}r^{a_t,t}] , \bar{O}^{\prime}) \\
& = \min_d\max_iE[\sum_{t}r^{i,t}_d - \sum_{t=1}^Tr^{a_t,t}_d].
\end{align*}


We use the Hoeffding's inequality, which leads to 
\begin{align} \label{eq:prob}
    P(C_A) & = P(\forall i: |\mu^{i}_d - \frac{\sum_{t }r_d^{i,t}}{N_i(T)}| \leq |\frac{K^{4}\log{\log{N_i(T)}}}{N_i(T)}|) \notag \\
    & \geq 1 - \eta
\end{align}
where $\eta = \sum_i 2\exp{\{-2N_i(T)\cdot(\frac{K^{{4}}\log{\log{N_i(T)}}}{N_i(T)})^2\}} = \sum_i 2\exp{\{-2\cdot\frac{(K^{4}\log{\log{N_i(T)}})^2}{N_i(T)}\}}$. 

Through the end of the proof, we assume that $K = 2$ and $11 <T \leq 1200$, i.e. 2-arm bandits with mean values $\mu^1,\mu^2$.

We first argue that $f(x) = \frac{(\log \log x)^2}{x}$ is decreasing for $x \geq 11$. We get $f^{\prime}(x) = \frac{\log \log x}{x^2} \cdot (\frac{2 - (\log \log x)\log x}{\log x})$. For $x \geq 11$, $\log x \geq 0$, $\log \log x \geq 0$ and both are increasing. We conclude that $(\log \log x) \cdot \log x$ is increasing for $x \geq 11$. Since $f^{\prime}(11) < 0$, it then follows that $f^{\prime}(x) < 0$ for $x \geq 11$. 

If there exists arm $i$ such that $N_i(T) = 1$, then we have $N_{j}(T) = T-1$ for $j \neq i $ since $K=2$. 

In this case $\frac{(K^{4}\log{\log{N_i(T)}})^2}{N_i(T)} = \infty$, which implies that
\begin{align*}
    \eta & = 2\exp{\{-\infty\}} +  2\exp{\{-2\frac{(K^{4}\log{\log{(T-1)}})^2}{T-1}\}} \\
    & = 2\exp{\{-2\frac{(K^{4}\log{\log{(T-1)}})^2}{T-1}\}} \\
    & \leq 2\exp{\{-2\frac{(K^{4}\log{\log{T}})^2}{T}\}} \\
    & = 2\exp{\{-2^{9}\frac{(\log{\log{T}})^2}{T}\}} \\
    & \leq 2 \cdot \exp{(-2^9 \cdot \frac{(\log \log 1200)^2}{1200})} \leq 2 \cdot \frac{1}{5} = \frac{2}{5}
\end{align*}
where the first inequality is by monotonicity of $\frac{(\log{\log{x}})^2}{x}$ for $x \geq 11$.

Suppose now that for $i = 1,2$ we have $N_i(T) \geq 2$, and $N_i(T) \leq T$. For $x = 2,3, \ldots, 1200$, $f(x)$ can only have maximum in $x \in \{2,3, \ldots, 10,1200\}$. Its maximum is for $x = 3$ and then the maximum in the $\eta$ term is at $N_i(T) = 3$. 

We thus have
\begin{align*}
    \eta & \leq 2 \cdot  2\exp{\{-2^{9}\frac{(\log{\log{3}})^2}{3}\}} < \frac{8}{9}.
\end{align*}

Therefore, we have $1-\eta  \geq \frac{1}{9} >0$. With probability $1-\eta$,
\begin{align*}
    R_T^{\prime} & = \min_d\max_i (\sum_tr^{i,t}_d - \sum_{j=1}^K\sum_{a_t=j}r^{j,t}_d) \\
    & \geq \min_d\max_i (T\cdot \mu_d^i - K^{4}\log{\log{T}}- \\
    & \qquad \qquad  \sum_{t=1}^T\mu^{a_t,t}_d - \sum_{j=1}^K|K^{4}\log{\log{Nj(T)}}|) \\
    & \geq \min_d\max_i (T\cdot \mu_d^i -  \sum_{t=1}^T\mu^{a_t,t}_d - (K+1)K^{4}\cdot \log{\log{T}}) \\
    & = \min_d{\bar{R}_T^{d}} - (K+1)K^{4}\log{\log{T}} \\
    & = \min_d{\bar{R}_T^{d}} - 48\log{\log{T}}
\end{align*}
where the first inequality is straightforward from (\ref{eq:prob}) and the second inequality holds by noticing $N_i(T) \leq T$. 

Consider the instance-dependent lower bound with $\theta = \mu^2_1-\mu^1_1 = \mu^2_2-\mu^1_2 > 0$. By the result in~\citep{lattimore_szepesvari_2020:ch16} for MAB, we have that for any $T$ and any algorithm with $\sqrt{T}$ regret upper bound, it must hold
\begin{align*}
    \bar{R}_T^{d} \geq \frac{\log{T}}{\theta}. 
\end{align*}

When choosing $\theta \leq \frac{1}{48}$, we derive 
\begin{align}\label{r_t}
    R_T^{\prime} & \geq \bar{R}_T^{d} - 48\log{\log{T}}  \\
    & \geq 48(\log{T} - \log{\log{T}}).\label{r_t}
\end{align}

To conclude, we obtain
\begin{align*}
    E[R_T^{\prime}] & \geq E[R_T^{\prime} \cdot I_{C_A}] \\
    & \geq 48(\log{T} - \log{\log{T}})\cdot(1-\eta) \\ 
    & \geq \frac{48}{9}O^*(\log{T}) = O^*(\log{T})
\end{align*}
where the second inequality holds by~(\ref{r_t}) and the last
inequality uses $1 - \eta > \frac{1}{9}$.


\end{proof}

\subsubsection{Proof of Theorem~\ref{th:pseudo_adv}}

\begin{proof}

\citep{bubeck2012regret} analyze the lower bound on pseudo regret for MAB and establish that for any $\epsilon >0 $, there exists an adversarial MAB, denoted by $(r_1^{i,t})$, satisfying
\begin{align}\label{eq:adv_r_p}
    \inf \bar{R}_T^1 \geq O^*(\sqrt{T}) - \epsilon
\end{align}
where $\inf$ is taken over all algorithms.

Again, we focus on adversarial MO-MAB with reward vectors constant in dimensions, i.e. $r^{i,t} = (r_1^{i,t}, \ldots, r_1^{i,t})$, which is equivalent to MAB in the sense that $\bar{R}_T^{\prime} = \bar{R}_T^1 = \ldots = \bar{R}_T^{D}$ given by
\begin{align*}
    \bar{O}^{\prime}_A = \{i^*\} = \argmax_iE[\sum_{t=1}^Tr_d^{i,t}]
\end{align*}
and subsequently by Lemma~\ref{lemma:2}, we have 
\begin{align*}
    \bar{R}^{\prime}_T & = Dist\left(E[\sum_{t=1}^Tr^{a_t,t}], \bar{O}^{\prime}\right) \\
    & = Dist\left(E[\sum_{t=1}^Tr^{a_t,t}], \{E[\sum_{t=1}^Tr_d^{i^{*},t}]\}\right) \\
    & = E[\sum_{t=1}^Tr_d^{i^*,t}] - E[\sum_{t=1}^Tr_d^{a_t,t}] = \bar{R}_T^d
\end{align*}
for any $ 1 \leq d \leq D$.

This is to say that $\inf \bar{R}_T^{\prime} \geq O^*(\sqrt{T})$ by (\ref{eq:adv_r_p}) holds for all algorithms.

\end{proof}

\subsubsection{Proof of Theorem~\ref{th:pareto_adv}}

\begin{proof}

Consider adversarial MO-MAB with reward vectors being $r^{i,t} = (r_1^{i,t}, \ldots, r_1^{i,t})$. The Pareto optimal set $O^{\prime}_A$ is the unique best arm $i^*$ that satisfies $i^* = \argmax_i\sum_{t=1}^Tr_1^{i,t}$. Then the MO-MAB problem essentially degenerates to MAB since $R_T^{\prime} = R_T^1 = \ldots = R_T^{D}$ which is guaranteed by 
\begin{align*}
    R^{\prime}_T & = Dist\left(\sum_{t=1}^Tr^{a_t,t}, O^{\prime}\right) \\
    & = \sum_{t=1}^Tr_d^{i^*,t} - \sum_{t=1}^Tr_d^{a_t,t} = R_T^d, \text{ for every } d.
\end{align*}

By the result in~\citep{gerchinovitz2016refined} stating that 
\begin{align*}
    R_T^1 \geq O^*(\sqrt{T})
\end{align*}
holds for any randomized algorithm, we have that the Pareto regret $R^{\prime}_T$ is larger than $\sqrt{T}$. 

\end{proof}

\subsubsection{Results in Section 5.2}
\subsubsection{Proof of Theorem~\ref{th:non-target}}

\begin{proof}[Proof of Theorem~\ref{th:non-target}]

With $\beta(n) = \sqrt{\frac{2\sigma^2}{n}\log{\frac{\pi^2Kn^2}{3\delta}}}$ where $\sigma^2$ is the variance of values for different dimensions in reward vectors of different arms, let us define event $E$ as
\begin{align*}
    E = \{\forall i, \forall t > K: ||\hat{\mu}^i(t) - \mu^i||_{\infty} < \beta(N_i(t))\}.
\end{align*}

We first present several lemmas used later. As before, without loss of generality we assume that in the first $K$ steps every arm is pulled. This implies $N_i(t) \geq 1$ for any $t > K$ and arm $i$. 

\begin{lemma}\label{le:E}
For any $\delta \in (0,1)$, $P(E) > 1-D\delta$.
\end{lemma}

\begin{proof}

Note that 
\begin{align}\label{eq:CH}
   & P(E) \notag \\
   & = P(\forall i, \forall t > K: ||\hat{\mu}^i(t) - \mu^i||_{\infty} < \beta(N_i(t))) \notag \\
   & = P(\forall i, \forall t > K, \forall d: \notag \\
   & \qquad \qquad |N_i^{-1}(t)\sum_{a_s = i}r_d^{i,s} - \mu_d^i| < \beta(N_i(t))) \notag \\
   & \geq 1 - 2\sum_{t>K}\sum_{i\leq K}D\exp{\{-\frac{N_i(t)}{2\sigma^2}\beta(N_i(t))^2\}} \notag \\
   & = 1- 2\sum_{t>K}\sum_{i \leq K}D\exp{\{-\frac{N_i(t)}{2\sigma^2}\frac{2\sigma^2}{N_i(t)}\log{\frac{\pi^2KN_i^2(t)}{3\delta}}\}} \notag \\
   & = 1 - 2\sum_{t>K}\sum_{i\leq K}D\left(\frac{\pi^2KN_i^2(t)}{3\delta}\right)^{-1} \notag \\
   & = 1 - \frac{6\delta}{\pi^2K}D\sum_{t>K}\sum_{i\leq K}N_i^{-2}(t) \notag\\
   & \geq  1 - D\delta \cdot \frac{6}{\pi^2K} \frac{K\pi^2}{6} \notag \\
   & = 1-D\delta 
\end{align}
where the first inequality is by the Hoeffding's inequality for sub-Gaussian distributions and the last inequality holds by the fact that $\sum_{t=1}^{\infty}\frac{1}{t^{-2}} = \frac{\pi^2}{6}$. 

\end{proof}

\begin{lemma}\label{le:N}
Let us assume event $E$ holds. Then, for any $i < K$ and any $t \geq 2K$, we have 
\begin{align*}
    N_i(t) \leq \min{\{N_K(t), 2 + \frac{9\sigma^2}{\Delta_0^2}\log{t}\}}.
\end{align*}

\end{lemma}

\begin{proof}

Since $t>2K$, if $N_j(t) \leq 2$ for any $j <K$, then $N_K(t) \geq 2$ and the result holds trivially. Let $S = \{j < K: N_j(t) >2 \} \neq \emptyset$. If we prove the statement for each arm in $S$, this implies $N_K(t) >2$ and thus the statement holds also for any arm $j \not\in S$. Let now $i \in S$, i.e. $N_i(t) > 2$ . 

If $a_t \neq i$, then we can consider the last time $t^{\prime}$ we had $a_{t^{\prime}} = i$ and apply the result in this case.

Let us assume $a_t = i$ and we consider the previous time step $t^{\prime} < t$ when Bob pulled arm $i$. Since $a_t = i$, it implies $i \in O_A^t$. We have that 
$N_i(t^{\prime}-1) + 1 = N_i(t^{\prime}) = N_i(t) - 1$, $N_i(t-1) = N_i(t^{\prime})$ by the definition of $N_i(t)$. We note that by definition of $\alpha_{t^{\prime}}$ we have $\bar{z}^j - \hat{z}^j + \frac{\alpha_{t^{\prime}}}{N_j(t^{\prime})} \geq 0$ for any $j \in O_A^{t^{\prime}}$. This implies since $i \in O_A^{t^{\prime}}$ and $\hat{z}^{i} - \frac{\alpha_{t^{\prime}}}{N_i(t^{\prime})} = \hat{\Tilde{\mu}}^i(t^{\prime})$, that 
\begin{align}\label{eq:attack_rule}
    \hat{\Tilde{\mu}}^{i}(t) \prec \hat{\Tilde{\mu}}^K(t) - (2\beta(N_{K}(t)) + \Delta_0) \cdot 1.
\end{align}

Since $a_t = i$ is chosen by Bob based on Pareto UCB, there exists at least one dimension $d$, such that 
\begin{align*}
   & \hat{\Tilde{\mu}}^{i}_d(t-1) + 3\sigma\sqrt{\frac{\log{t}}{N_i(t-1)}} \\ &
   \geq \hat{\Tilde{\mu}}^{K}_d(t-1) 
    + 3\sigma\sqrt{\frac{\log{t}}{N_K(t-1)}} \\
   & \text{and thus } \\ & \hat{\Tilde{\mu}}^{i}_d(t^{\prime})+ 3\sigma\sqrt{\frac{\log{t}}{N_i(t^{\prime})}} \\
   & \qquad \quad \geq \hat{\Tilde{\mu}}^{K}_d(t-1) + 3\sigma\sqrt{\frac{\log{t}}{N_K(t-1)}} \\
   & \text{which is equivalent to} \\
   & \qquad 3\sigma\sqrt{\frac{\log{t}}{N_i(t^{\prime})}} - 3\sigma\sqrt{\frac{\log{t}}{N_K(t-1)}} \geq \hat{\Tilde{\mu}}^{K}_d(t-1)  - \hat{\Tilde{\mu}}^{i}_d(t^{\prime}).
\end{align*}
We have
\begin{align*}
    & \hat{\Tilde{\mu}}^{K}_d(t-1)  - \hat{\Tilde{\mu}}^{i}_d(t^{\prime}) \\
    & \geq \hat{\mu}^{K}_d(t-1) - \hat{\mu}^{K}_d(t^{\prime}) + 2\beta(N_K(t^{\prime})) + \Delta_0 \\
    & \geq -\beta(N_K(t-1)) - \beta(N_K(t^{\prime})) + 2\beta(N_K(t^{\prime})) + \Delta_0 \\
    & \geq \Delta_0 > 0.
\end{align*}
The first inequality is due to (\ref{eq:attack_rule}) and since Alice never attacks arm $K$ we have $\hat{\Tilde{\mu}}^K(t-1) = \hat{\mu}^K(t-1)$. The second inequality holds by the definition of event $E$ and the third inequality uses the fact that $\beta(n) = \sqrt{\frac{2\sigma^2}{n}\log{\frac{\pi^2Kn^2}{3\delta}}}$ is monotone decreasing in $n \geq 1$ if $K \geq \frac{3\delta e^2}{\pi^2}$ which can be shown by calculus.

Therefore, we have that 
\begin{align}\label{eq:i_K}
 & 3\sigma\sqrt{\frac{\log{t}}{N_i(t^{\prime})}} - 3\sigma\sqrt{\frac{\log{t}}{N_K(t-1)}} \geq \Delta_0 > 0, \\
 & \text{and thus } N_K(t) = N_K(t-1) \geq N_i(t^{\prime}) + 1 = N_i(t). \notag
\end{align}

Meanwhile, from (\ref{eq:i_K}) we get 
\begin{align*}
    & 3\sigma\sqrt{\frac{\log{t}}{N_i(t^{\prime})}} > \Delta_0 \\
    & \text{ and thus }\\
    & N_i(t) = 1+ N_i(t^{\prime}) \leq 2 + \frac{9\sigma^2}{\Delta_0^2}\log{t},
\end{align*}
which completes the proof.

\end{proof}

\begin{lemma}\label{le:cost}
Let us assume event $E$ holds. Then the cumulative attack cost for any arm $i < K$ up to time step $t \geq 2K$, can be bounded as
\begin{align*}
    \sum_{\substack{a_s  = i \\ s \leq t}}\alpha_s \leq \max_j N_j(t)(\Delta_j + \Delta_0 + 4\beta(N_j(t))).
\end{align*}
\end{lemma}

\begin{proof}
Note that the right hand side is monotone increasing in $t$. It suffices to show that the result holds for $t$ with $a_t = i$ essentially assuming that $N_K(t-1) = N_K(t)$.

By its definition, the cumulative cost satisfies

\begin{align*}
   \alpha_t & =  \max \{ \max_{j \in O_A^t, d} N_j(t)(\hat{z}^j_d - \bar{z}^j_d),0\}\\
    & = \max \{\max_{j \in O_A^t,d} N_j(t)\hat{\mu}_d^j(t) - \sum_{s=1}^{t-1}\alpha_s - N_j(t) \cdot (\hat{\mu}^K_d(t) - \\
    & \qquad 2\beta(N_{K}(t)) + \Delta_0) ,0\} \\
    & = \max \{ \max_{j \in O_A^t,d}(N_j(t))(\hat{\mu}^j_d(t) -(\hat{\mu}_d^K(t) - \\
    & \qquad 2\beta(N_{K}(t)) - \Delta_0))  - \sum_{\substack{s=1} }^{t-1}\alpha_s  ,0\},
\end{align*}
which implies by adding $\sum\limits_{\substack{s=1 \\ a_s = i}}^{t-1}\alpha_s$ to both sides that 
\begin{align}\label{eq:cost_sum}
    \sum_{\substack{s=1 \\ a_s = i}}^{t}\alpha_s 
    & \leq \max \{ \max_{j \in O_A^t, d}  \{ N_j(t)(\hat{\mu}^{j}_d(t) - \hat{\mu}^K_d(t-1) + \notag \\
    & \qquad 2\beta(N_K(t-1)) + \Delta_0)\},\sum_{\substack{s=1 \\ a_s = i}}^{t-1}\alpha_s\}.
\end{align}
Note that $\sum\limits_{\substack{s=1 \\ a_s = i}}^{t-1}\alpha_s \leq \sum\limits_{\substack{s=1}}^{t-1}\alpha_s$. Since event $E$ holds, we have that 
\begin{align*}
    & \sum\limits_{\substack{s=1 \\ a_s = i}}^{t}\alpha_s \\ 
    & \leq \max \{   \max_{j \in O_A^t,d} N_j(t)(\mu^j_d + \beta(N_j(t))) - \mu^K_d +  \\
    & \qquad \qquad \beta(N_K(t-1)) + 2\beta(N_K(t-1)) + \Delta_0, \sum\limits_{\substack{s=1 \\ a_s = i}}^{t-1}\alpha_s\} \\
    & \leq \max \{ \max_{j \in O_A^t,d} N_j(t)(\mu^j_d- \mu^K_d + 4\beta(N_j(t))) + \Delta_0,\sum\limits_{\substack{s=1 \\ a_s = i}}^{t-1}\alpha_s\} \\
    & = \max \{\max_{j \in O_A^t} N_j(t)(\Delta_j + \Delta_0 + 4\beta(N_j(t))),\sum\limits_{\substack{s=1 \\ a_s = i}}^{t-1}\alpha_s\} \\
    & \leq  \max_{j \in O_A^t} N_j(t)(\Delta_j + \Delta_0 + 4\beta(N_j(t)))
\end{align*}
where the first inequality holds by the definition of event $E$, the second inequality holds as a result of Lemma~\ref{le:N} and the monotonicity of $\beta(\cdot)$ and the third inequality holds by the induction over $\sum_{s=1}^{t-1}\alpha_s$. The induction is with respect to all times $\bar{t}$ such that $a_{\bar{t}} = i$. The base case corresponds to $1 \leq \bar{t} \leq 2K$ since we assume that initially every arm is pulled. If there exists $\bar{t}$, $K \leq \bar{t}  \leq 2K$, such that $a_{\bar{t}} = i$, then this is the base case. Otherwise $1 \leq \bar{t} < K$ with $a_t = i$ and this is the base case. We also use the fact that $\sum_{s=1}^{2K}\alpha_s = 0$.

\end{proof}

Suppose event $E$ holds. Lemma~\ref{le:N} proves the first half of the theorem. 

For the second half, we observe that the total attack cost satisfies 
\begin{align*}
    & \quad \sum_{i < K}\sum\limits_{\substack{a_s = i \\ s \leq T}}\alpha_s \\
    & \leq \sum_{i < K} \max_j N_j(T)(\Delta_j + \Delta_0 + 4\beta(N_j(T))) \\
    & \leq (K-1)\left(2 + \frac{9\sigma^2}{\Delta_0^2}\log{T}\right) \cdot \\
    & \qquad \max_j(\Delta_j + \Delta_0 + 4\beta(N_j(t))) \\
    & \leq (K-1)\left(2 + \frac{9\sigma^2}{\Delta_0^2}\log{T}\right)\max_j(\Delta_j + \Delta_0 + 4\beta(2))  \\
    & = (K-1)\left(2 + \frac{9\sigma^2}{\Delta_0^2}\log{T}\right)\max_j(\Delta_j + \Delta_0) + O^*(\log{T}) 
\end{align*}
where the first inequality holds by Lemma~\ref{le:cost} and the second inequality again uses Lemma~\ref{le:N}. The last inequality holds by the fact that $\beta(\cdot)$ is monotone decreasing.

According to Lemma~\ref{le:E}, we have $P(E) \geq 1-D\delta$ and subsequently the results hold with probability at least $1-D\delta$.

\end{proof}

\subsubsection{Proof of Theorem~\ref{th:Pareto-regret}}

\begin{proof}[Proof of Theorem~\ref{th:Pareto-regret}]

We note that 
\begin{align*}
  R_T & = \sum_{i=1}^{K}N_i(T)Dist(\mu^{i}, O) \\
      & \geq N_K(T) \cdot Dist(\mu^K, O).
\end{align*}

By construction of the instances aforementioned, for any arm $i \in O$, we have $\mu^K \prec \mu^i$ which leads to $Dist(\mu^K, O) > 0$. 

By Theorem~\ref{th:non-target} we derive $N_K(T) = T - \sum_{i<K}N_i(T) \geq T - (K-1)O^*(\log T) = O^*(T)$.

To conclude, the Pareto regret satisfies $R_T \geq N_K(t) \cdot Dist(\mu^K, O) = O^*(T)$ since $\mu^K \prec \mu^i$ implies $Dist(\mu^K,O) > 0$.\end{proof}

\subsubsection{Proof of Theorem~\ref{th:Pareto-regret-new-1}}

\begin{proof}

The proof is two-fold. We first show that the two Pareto optimal fronts $\bar{O}^{\prime}$ and $O^{\prime}$ can be quite close in a high probability sense and then show that the obtained reward vectors approach the mean vectors as $T$ goes large by concentration inequalities. 

We first note that the conditions in the theorem imply $\gamma \leq \frac{1}{5}$ and $K = 2$ is valid for Theorem 9. This implies that we can meet Assumption 4. 

Let $\eta$ be fixed with $0 < \eta < 1$. Since $\{r^{i,t}\}_{1 \leq t \leq T}$ are i.i.d. sub-Gaussian distributed, the Chernoff-Hoeffding inequality implies
\begin{align}\label{eq:concen_ineq}
    & P(|\frac{1}{N_i(T)} \cdot \sum_{a_s = i} r^{i,s} - \mu^{i}| \not \preceq \gamma \cdot 1) \leq 2D\exp{(-\frac{\gamma^2}{2\sigma^2})}.
\end{align}
In (\ref{eq:concen_ineq}) we use the derivation from (\ref{eq:CH}). 

By choosing $\gamma$ such that $2DK\exp{(-\frac{\gamma^2}{2\sigma^2})} \leq \eta$ which is imposed in the theorem, we have 
\begin{align}\label{eq:CHnew}
    & P(|\frac{1}{N_i(T)} \cdot \sum_{a_s = i} r^{i,s} - \mu^{i}| \preceq \gamma \cdot 1, \forall 1 \leq i \leq K) \notag \\
    & \geq 1 - \eta.
\end{align}
We denote the event as $E_0$ when using (\ref{eq:CHnew}), i.e. $P(E_0) \geq 1-\eta$.

Likewise, note that
\begin{align}\label{eq:concen_ineq_T}
    & P(|\frac{1}{T} \cdot \sum_{t=1}^T r^{i,t} - \mu^{i}| \preceq \gamma \cdot 1, \forall 1 \leq i \leq K) \geq 1-\eta
\end{align}
which can be shown by
\begin{align*}
    & P(|\frac{1}{T} \cdot \sum_{t=1}^T r^{i,t} - \mu^{i}| \preceq \gamma \cdot 1, \forall 1 \leq i \leq K) \\
    & = P(\forall i, \forall d: |\frac{1}{T} \cdot \sum_{t=1}^T r^{i,t}_d - \mu_d^i| < \gamma) \\
   & = 1 - P(\exists i, \exists d: |\frac{1}{T} \cdot \sum_{t=1}^T r^{i,t}_d - \mu_d^i| > \gamma) \\
   & \geq 1 - \sum_{i=1}^K\sum_{d =1}^D P(|\frac{1}{T} \cdot \sum_{t=1}^T r^{i,t}_d - \mu^i_d| > \gamma) \\ 
   & \geq 1 - 2DK\exp{(-\frac{T\gamma^2}{2\sigma^2})} \\
   & \geq 1 - \eta
\end{align*}
where the first inequality is by the Bonferroni's inequality, the second inequality is by the Chernoff-Hoeffding inequality and the last one holds by the choice of $\gamma$. Note that if we use (\ref{eq:concen_ineq_T}), we denote the event as $E_1$, i.e. $P(E_1) \geq 1-\eta$.

Meanwhile, by Assumption 4, $\mu^j - \mu^i \succeq 5\gamma \cdot 1$
holds for any arm $i$ not in $\bar{O}_A$ and arm $j$ in $\bar{O}_A$. 

Note that $O$ and $\bar{O}^{\prime}$ only differs in whether $\frac{1}{N_i(T)}\sum_{a_t = i}\alpha_t$ is present. Since $K=2$, the two terms for attack cost are equivalent by noting that
\begin{align}\label{eq:K=2}
\frac{1}{N_i(T)}\sum_{a_t = i}\alpha_t = \frac{1}{\sum_{j \neq K}N_j(T)} \cdot \sum_{a_t \neq K}\alpha_t.
\end{align}

For any arm $j \in \bar{O}^{\prime}_A$ and arm $i$ not in $\bar{O}^{\prime}_A$, on event $E_1$ we derive
\begin{align}\label{O_O}
& -\frac{1}{T} \cdot \sum_{t=1}^T r^{i,t} + \frac{1}{T} \cdot \sum_{t=1}^T r^{j,t}  \notag \\
& = -\frac{1}{T} \cdot \sum_{t=1}^T r^{i,t} + \mu^i- \mu^i + \notag \\
& \qquad \frac{1}{T} \cdot \sum_{t=1}^T r^{j,t} - \mu^j + \mu^j \notag \\
&  = -\mu^i+\mu^j - \notag \\
& \qquad  (\frac{1}{T} \cdot \sum_{t=1}^T r^{i,t} - \mu^{i}) + (\frac{1}{T} \cdot \sum_{t=1}^T r^{j,t} - \mu^{j}) \notag \\
& \succeq (5\gamma- 2\gamma) \cdot 1 = 3\gamma \cdot 1> 0.
\end{align}
where the last inequality is by the choice of $i,j$ and (\ref{eq:concen_ineq_T}).

This implies that $i \not \in O_A^{\prime}$ together with (\ref{eq:K=2}).

Similarly, for any arm $j \in O^{\prime}_A$ and arm $i \not \in O^{\prime}_A$, on event $E_1$ we have 
\begin{align}\label{O_O_new}
& -\mu^j +\mu^i \notag \\
&  = -\frac{1}{T} \cdot \sum_{t=1}^T r^{j,t} + \frac{1}{T} \cdot \sum_{t=1}^T r^{i,t}  - \notag \\
& \qquad  (\frac{1}{T} \cdot \sum_{t=1}^T r^{i,t} - \mu^{i}) + (\frac{1}{T} \cdot \sum_{t=1}^T r^{j,t} - \mu^{j}) \notag \\
& \preceq (0 + \gamma + \gamma) \cdot 1 = 2\gamma \cdot 1
\end{align}
where the last inequality is by the choice of $i,j$ and (\ref{eq:concen_ineq_T}). Note that since $K =2$ we have $\sum_{t=1}^T r^{i,t} \preceq \sum_{t=1}^T r^{j,t}$.

If $i \in \bar{O}_A^{\prime}$, then $j \not \in \bar{O}_A^{\prime}$ and $\mu^i > \mu^j + 5\gamma \cdot 1$ since 1) there is an arm not in $\bar{O}_A^{\prime}$ (by our construction $\bar{O}_A^{\prime}$ contains at least one arm, arm $K$), and 2) Assumption 4 is valid. This contradicts (\ref{O_O_new}) and thus implies that $i \not \in \bar{O}_A^{\prime}$.

Consequently, we established that on event $E_1$ we have $\bar{O}_A^{\prime} = O^{\prime}_A$.

Since $P(E_1) \geq 1-\eta$ we conclude
\begin{align}\label{eq:O_eq}
 P( \bar{O}^{\prime}_A = O^{\prime}_A ) \geq 1 - \eta.
\end{align}

We further obtain by (\ref{eq:concen_ineq_T}) and (\ref{eq:K=2}), on event $E_1$ for any $i$, 
\begin{align}\label{eq:O_O_P}
    & (\frac{1}{T}\sum_{t=1}^T r^{i,t} - \frac{1}{N_i(T)}\sum_{a_t = i}\alpha_t \cdot 1) - \gamma \notag \\
    & \qquad \preceq \mu^{i} - \frac{1}{\sum_{j \neq K}N_j(T)} \cdot \sum_{a_t \neq K}\alpha_t \cdot 1  \notag \\
    & \qquad  \preceq (\frac{1}{T}\sum_{t=1}^T r^{i,t} - \frac{1}{N_i(T)}\sum_{a_t = i}\alpha_t \cdot 1) + \gamma.
\end{align}

Therefore, by the definition of $\bar{O}^{\prime}$ being defined as the Pareto optimal front of vectors $\mu^{i} - \frac{1}{\sum_{j \neq K}N_j(T)} \cdot \sum_{a_t \neq K}\alpha_t \cdot 1$ over all arm $i$ and $O^{\prime}$ being over vectors $\frac{1}{T}\sum_{t=1}^T r^{i,t} - \frac{1}{N_i(T)}\sum_{a_t = i}\alpha_t \cdot 1$, we have on event $E_1$ that 
\begin{align}\label{eq:R_T_1}
    & R^{\prime}_T \notag \\
    & = T \cdot  Dist(\frac{1}{T}\sum_{t=1}^Tr^{a_t,t} \notag \\ 
    & \qquad \qquad - \frac{1}{\sum_{i \neq K} N_i(T)}\sum_{a_t \neq K}\alpha_t \cdot 1, O^{\prime})  \notag \\
    & \geq T \cdot  Dist(\frac{1}{T}(N_K(T) \cdot \hat{\mu}^{K} + \sum_{i \neq K} N_i(T) \cdot \hat{\mu}^{i}) \notag \\ 
    & \qquad \qquad - \frac{1}{\sum_{i \neq K} N_i(T)}\sum_{a_t \neq K}\alpha_t \cdot 1, \bar{O}^{\prime}) - 2\gamma T.
\end{align}
The inequality follows from (\ref{eq:O_eq}) and (\ref{eq:O_O_P}) and the following statement in Proposition 1. 

\paragraph{Proposition 1} For any $\gamma_1 > 0$, $\gamma_2 > 0$ if $O_A^{\prime} = \bar{O}_A^{\prime}$ and $u^a \succeq v^a - \gamma_1 \cdot 1$ for any $v^a \in \bar{O}^{\prime},u^a \in O^{\prime}$ where $a \in O_A^{\prime}$, then for any $e$,$\bar{e}$ with $e \preceq \bar{e} + \gamma_2 \cdot 1$ by Lemma~\ref{lemma:1} we have 
\begin{align*}
    Dist(e, O^{\prime}) & = \min_d\max_{a \in O_A^{\prime}}{(u^a_d - e_d)} \\
    & = \min_d\max_{a \in \bar{O}_A^{\prime}}{(u^a_d - e_d)} \\
    & \geq \min_d\max_{a \in \bar{O}_A^{\prime}}{(v^a_d - e_d - \gamma_1)} \\
    & \geq \min_d\max_{a \in \bar{O}_A^{\prime}}{(v^a_d - \bar{e}_d )} - \gamma_1 - \gamma_2 \\
    & = Dist(\bar{e},\bar{O}^{\prime}) - \gamma_1 - \gamma_2.
\end{align*}\qed

Note that $Dist(e,A) = Dist(e-z,\{a -z | a\in A\})$. On event $E_0 \cap E_1$, we have that 
\begin{align*}
 & R_T^{\prime}  \\
 & \geq T \cdot Dist\left(\frac{1}{T} \cdot \left( N_K(T)\cdot \hat{\mu}^{K} + \sum_{i \neq K} N_i(T) \cdot \hat{\mu}^{i}\right), O\right)  \\
 & \qquad \qquad - 2\gamma T \\
 & \geq T \cdot Dist\left(\frac{1}{T} \cdot \left( N_K(T)\cdot \mu^{K} + \sum_{i \neq K} N_i(T) \cdot \mu^{i}\right), O\right) \\
  & \qquad \qquad - 3\gamma T \\
  & \doteq  T\mathcal{D}  - 3\gamma T.
\end{align*}
where in the first inequality we use (\ref{eq:K=2}), the various definitions and the second inequality is a result of ($\ref{eq:CHnew}$) by bounding $\mu_i$ with $\hat{\mu} + \gamma \cdot 1$ for any arm $i$.

By Theorem~\ref{th:non-target}, we have that on event $E$
\begin{align}\label{Fact_N}
N_i(T) \leq O(log T)
\end{align}
and clearly $N_K(T) \leq T$.

Therefore, on events $E,E_0,E_1$ we have
\begin{align} \label{eq:D_ineq}
T\mathcal{D}
& \geq T \cdot Dist(\frac{1}{T} \cdot (T \cdot \mu^{K} + \notag \\
& \qquad (K-1)O(logT) \max_{i \neq K}\mu^i) , O) \notag \\
& \geq T \cdot Dist(\mu^K, O) - (K-1)O(logT)
\end{align}
where the last inequality uses the fact that $\max_i \mu^i  \leq 1 $.

By our construction, the distance from the reward vector of arm $K$ to the Pareto optimal front $O$, $Dist(\mu^K, O) \geq 5\gamma$ since $K \not \in O$. Therefore, on $E \cap E_0 \cap E_1$ we have that
\begin{align*}
    R_T^{\prime} & \geq T\mathcal{D} - 3\gamma T \\
    & \geq T \cdot 5\gamma - (K-1)O(logT)  - 3\gamma T \\
    & = 2\gamma T - (K-1) \cdot O(logT).
\end{align*}
Note that $P(E \cap  E_0 \cap E_1) \geq 1 - P(E^c) - P((E_0 \cap E_1)^c) = - D\delta + P(E_0 \cap E_1) \geq - D\delta + 1 - P(E_0^c) - P(E_1^c) = 1 - D\delta - 2\eta$. This completes the proof.

\end{proof}

\subsubsection{Proof of Theorem~\ref{th:Pareto-regret-new-2}}

\begin{proof}
We use the notation from the proof of Theorem~\ref{th:Pareto-regret-new-1}. By the statement in~(\ref{eq:concen_ineq_T}), we obtain
\begin{align*}
    & P(|\frac{1}{T} \cdot \sum_{t=1}^T r^{i,t} - \mu^{i}| \preceq \gamma \cdot 1) \geq 1-\eta.
\end{align*}

By Assumption 4, $\mu^j - \mu^i \succeq 5\gamma \cdot 1$
holds for any arm $i$ not in $\bar{O}^{\prime}_A$ and arm $j$ in $\bar{O}^{\prime}_A$.

For any arm $j \in \bar{O}^{\prime}_A$ and arm $i$ not in $\bar{O}^{\prime}_A$, on event $E_1$ we have 
\begin{align*}
-\frac{1}{T} \cdot \sum_{t=1}^T r^{i,t} + \frac{1}{T} \cdot \sum_{t=1}^T r^{j,t} \succ 0 
\end{align*}
by the result in (\ref{O_O}).

Meanwhile, since $\bar{O}^{\prime}$ and $O^{\prime}$ consider the same attack cost, this again implies that $i \not \in O_A^{\prime}$.

Similarly, for any arm $j \in O^{\prime}_A$ and arm $i \not \in O^{\prime}_A$, on event $E_1$ we have arm $i \not \in \bar{O}^{\prime}_A$.

Consequently on event $E_1$ we get
\begin{align*}
    O^{\prime}_A =  \bar{O}^{\prime}_A 
\end{align*}
and
\begin{align*}
    & (\frac{1}{T}\sum_{t=1}^T r^{i,t} - \frac{1}{T}\sum_{t =1}^T\bar{\alpha}^i_t \cdot 1) - \gamma \\
    & \qquad \preceq \mu^{i} - \frac{1}{T}\sum_{t =1}^T\bar{\alpha}^i_t \cdot 1   \\
    & \qquad  \preceq (\frac{1}{T}\sum_{t=1}^T r^{i,t} - \frac{1}{T}\sum_{t =1}^T\bar{\alpha}^i_t \cdot 1) + \gamma .
\end{align*}
Therefore, by the definition of $\bar{O}^{\prime}$ being defined as the Pareto front of vectors 
$\mu^i - \frac{1}{T} \sum_{t=1}^T\bar{\alpha}^i_t \cdot 1$, whereas $O^{\prime}$ is over $\frac{1}{T}(\sum_{t=1}^T r^{i,t} - \sum_{t =1}^T\bar{\alpha}^i_t \cdot 1)$, we have on event $E_0 \cap E_1$ that 
\begin{align} \label{eq:R_T_2}
    R^{\prime}_T 
    & \geq T \cdot Dist(\frac{1}{T} \cdot \left( N_K(T)\cdot \mu^{K} + \sum_{i \neq K} N_i(T) \cdot \mu^{i}\right) - \notag \\
    & \qquad \qquad \frac{1}{T} \sum_{t=1}^T\bar{\alpha}_t^{a_t} \cdot 1, \bar{O}^{\prime}) -  2\gamma T.
\end{align}
This derivation follows (\ref{eq:R_T_1}), ($\ref{eq:CHnew}$) and relies on Proposition 1.

Consider the counterfactual attack cost $\bar{\alpha}_t^i$ on arm $i$ in the definition of $\bar{O}^{\prime}$. Formally, according to the definition of $\bar{\alpha}_t^i$, it reads explicitly as
\begin{align*}
    \bar{\alpha}_t^i & =  \max \{ \max_{d} N_i(t)(\hat{z}^i_d - \bar{z}^i_d),0\}.
\end{align*}
We observe that 
\begin{align*}
    \alpha_t &  = \max \{ \max_{j \in O_A^t, d} N_j(t)(\hat{z}^j_d - \bar{z}^j_d),0\} \\ 
    & \geq \bar{\alpha}_t^{a_t}
\end{align*}
since the chosen arm by Bob $a_t \in O_A^t$. 

Therefore, by the results in Lemma~\ref{le:cost} and Lemma~\ref{le:N} on event $E$, we have
\begin{align*}
   & \sum_{t=1}^T\bar{\alpha}_t^{a_t} \leq \sum_{t=1}^T\alpha_t \\
   & \leq (K-1)\max_{j < K}\sum_{\substack{a_s  = j \\ s \leq T}}\alpha_s \\
   & \leq (K-1)\max_j N_j(T)(\Delta_j + \Delta_0 + 4\beta(N_j(T))) \\ 
    & \leq (K-1)\left(2 + \frac{9\sigma^2}{\Delta_0^2}\log{T}\right)\max_j(\Delta_j + \Delta_0 + 4\beta(2))  \\
    & = O(\log{T}).
\end{align*}

This leads to $\frac{1}{T}\cdot \sum_{t=1}^T\bar{\alpha}_{t}^{a_t} \to 0$ as $T \to \infty$.

Meanwhile for any $i \neq K$, we have
\begin{align*}
   & \sum_{t=1}^T\bar{\alpha}_t^{i} \\
   & = \sum_{t=1}^T\max \{ \max_{d} N_i(t)(\hat{z}^i_d - \bar{z}^i_d),0\} \\
   & \leq \sum_{t=1}^T \max \{ \max_{j \in O_A^t,d} N_i(t)(\hat{z}^j_d - \bar{z}^j_d),0\} \\
   & = \sum_{t=1}^T\alpha_{t} \leq O(\log{T}).
\end{align*}

As a result, we further obtain $\frac{1}{T}\cdot \sum_{t=1}^T\bar{\alpha}_{t}^{i} \to 0$ as $T \to \infty$. 

Suppose for $T \geq T(\gamma)$ we have $ \frac{1}{T}\cdot \sum_{t=1}^T\alpha_{t}^{i} \leq \gamma$ and $\frac{1}{T}\cdot \sum_{t=1}^T\alpha_{t}^{a_t} \leq \gamma$.

From~(\ref{eq:R_T_2}) we further obtain by using Proposition 1
\begin{align*}
 & R_T^{\prime} \\
 & \geq T \cdot Dist\left(\frac{1}{T} \cdot \left( N_K(T)\cdot \mu^K + \sum_{i \neq K} N_i(T) \cdot \mu^i\right), O\right)\\
 & \qquad \qquad  - 2\zeta \cdot T   -2\gamma T \\ 
 & \doteq  T\mathcal{D} - 2\gamma T -  2\gamma T.
\end{align*}

By the result in (\ref{eq:D_ineq}),  we further have that on $E \cap E_0 \cap E_1$
\begin{align*} 
 T\mathcal{D} \geq T \cdot Dist(\mu^K, O) - (K-1) \cdot O(logT) 
\end{align*}

Since by assumption we have $Dist(\mu^K,O) \geq 5\gamma$, we conclude that for $T \geq T(\gamma)$
with probability at least $1-2\eta-D\delta$, $R^{\prime}_T$ is of order $T$ by noting that
\begin{align*}
    & R^{\prime}_T \geq T\mathcal{D} - 2\gamma T - 2\gamma T \\
    & \geq T \cdot 5\gamma - (K-1) \cdot O(logT) - 2 \gamma T -2\gamma T  \\
    & = \gamma T - (K-1) \cdot O(logT).
\end{align*}

\end{proof}

\end{document}